\documentclass[10pt]{article}
\PassOptionsToPackage{table}{xcolor}
\usepackage[preprint]{tmlr}

\usepackage{wrapfig,graphicx}
\usepackage{booktabs,longtable,multirow,makecell,array}
\usepackage{xcolor,colortbl}
\usepackage{amsmath,amssymb,siunitx}
\usepackage{tcolorbox}
\tcbuselibrary{skins,breakable}
\usepackage{pgfplots}
\pgfplotsset{compat=1.18}
\usepackage{enumitem,caption}

\definecolor{cohortbg}{RGB}{237,233,254}%
\definecolor{cohortfg}{RGB}{109,40,217}%
\definecolor{unipactfg}{RGB}{0,0,0}
\definecolor{checkgreen}{RGB}{22,163,74}%
\definecolor{crossgray}{RGB}{156,163,175}%
\definecolor{headerrow}{RGB}{249,250,251}%
\definecolor{notecolor}{RGB}{107,114,128}%
\definecolor{catgray}{gray}{0.95}
\definecolor{highlightgreen}{HTML}{E8F5E9}

\usepackage{amsmath,amsfonts,bm}

\def\eqref#1{equation~\ref{#1}}

\def\1{\bm{1}}

\DeclareMathAlphabet{\mathsfit}{\encodingdefault}{\sfdefault}{m}{sl}
\SetMathAlphabet{\mathsfit}{bold}{\encodingdefault}{\sfdefault}{bx}{n}

\usepackage{hyperref}
\hypersetup{hidelinks}
\usepackage{url}
\title{Retrieval-Augmented Multimodal Language Model with Heterogeneous Evidence Integration for Clinical Prognosis}
\author{\name Hamza Shafiq \email m.h.shafiq@tue.nl \\
      \addr Eindhoven University of Technology
      \AND
      \name Jialu Tang \email j.tang@tue.nl \\
      \addr Eindhoven University of Technology
      \AND
      \name Ting Dang \email Ting.Dang@unimelb.edu.au \\
      \addr University of Melbourne
      \AND
      \name Jun Hu \email j.hu@tue.nl \\
      \addr Eindhoven University of Technology
      \AND
      \name Aaqib Saeed \email a.saeed@tue.nl \\
      \addr Eindhoven University of Technology
}
\def\month{10}%
\def\year{2026}%
\def\openreview{\url{https://openreview.net/forum?id=5ChJTFLi5r}} 

\begin{document}
\maketitle
\begin{abstract}
Clinical prognosis requires integrating heterogeneous evidence from
structured measurements, physiological waveforms, longitudinal history,
and population context. Multimodal language models provide a natural
framework for this integration, but standard approaches often serialize
structured measurements as text, treat encounters independently, and lack
a mechanism for incorporating representations of clinically similar
patients. We introduce PRISM (Population-Retrieved Inference for
Structured Multimodal Prognosis), which represents heterogeneous evidence
through per-feature numerical tokenization, recency-weighted aggregation
of longitudinal ECGs, and a dual-modality population memory bank.
Retrieval operates in representation space: the model receives
mean-pooled embeddings of nearest training-cohort encounters, while no
individual retrieved record enters the prompt.
On the 1{,}443-task MDS-ED benchmark (MIMIC-IV), PRISM attains the
strongest reported figures across all four task groups: 86.2\% Diagnosis,
93.1\% Deterioration, 91.8\% ICU admission, and 92.8\% Mortality AUROC.
These figures use a 96-hour pre-arrival window that the reported
baselines did not receive. Under the benchmark input window alone, PRISM
reaches 0.8895 AUROC, compared with 0.8937 for UniPACT and 0.8889 for
MDS-ED.
A matched-input compact fusion model receiving the same five evidence
streams and frozen upstream encoders reaches 0.7652 AUROC, compared with
0.9092 for PRISM. Ablations show that replacing the numerical EHR
pathway with full text serialization reduces AUROC by 3.66 points,
removing the additional 96-hour pre-arrival history reduces it by 1.97
points, and zeroing retrieval reduces it by 0.99 points.
Random-neighbor controls show a 0.97-point macro-average advantage
for genuine retrieval; however, a paired task-level sign test does not
establish a uniform task-level advantage ($p=0.067$). On the external MC-MED cohort, supervised fine-tuning yields 0.8611 overall AUROC, compared with 0.8227 for locally trained gradient boosting and 0.7841 for source-only PRISM. These findings support heterogeneous evidence integration and supervised external adaptation.
\end{abstract}
\section{Introduction}
Clinical prognosis requires integrating multiple sources of patient information. Predicting whether a patient will deteriorate, require intensive care, or die in hospital demands the simultaneous interpretation of continuous physiological waveforms, dense panels of laboratory analytes and vital signs, longitudinal acquisition histories, and the implicit epidemiological priors that experienced clinicians carry from thousands of prior cases~\citep{smith2013news,raghunath2020prediction}. Large language models have emerged as a compelling backbone for this integration: their capacity for cross-modal attention over heterogeneous token sequences offers, in principle, a unified substrate for fusing structured and unstructured clinical data into coherent prognostic judgments~\citep{singhal2023large}. In practice, however, current multimodal clinical LLMs leave substantial predictive signal unrecovered due to three structural limitations in how they represent, contextualize, and reason over clinical evidence.

The most immediate of these concerns how structured clinical data enters the model. Systems that fuse waveform encoders with language model backbones~\citep{liu2023llava,li2023llavamed,tang2026unipact,pham2025qheart} typically rely on \emph{textualization}: laboratory values, vital signs, and demographics are serialized into natural-language strings and consumed through the tokenizer. This conversion does not explicitly preserve the ordinal and ratio-scale structure of numerical measurements at the input boundary. A troponin of 0.11\,ng/mL and a troponin of 0.01\,ng/mL become sequences of subword tokens whose clinical distance is opaque to the embedding layer~\citep{gorishniy2021revisiting}. Analytes such as troponin, lactate, and creatinine, among the strongest independent predictors of acute deterioration and in-hospital mortality~\citep{ahmed2014prognostic,kruse2011blood}, are either collapsed into coarse text summaries or discarded entirely. The resulting token sequence therefore carries the values without an explicit encoding of their numerical geometry.

A related limitation concerns the temporal dimension. Serial changes in a patient's ECG carry independent prognostic information beyond any single acquisition~\citep{lehmacher2020serial}: a new ST deviation against a previously normal baseline has different clinical meaning than the same morphology in a patient with chronic abnormalities. Yet current multimodal systems encode only the index-visit waveform~\citep{tang2026unipact,zhao2024ecgchat,pham2025qheart}, treating each encounter as if it were the patient's first and forfeiting the prognostic signal embedded in longitudinal change. This is a particularly costly omission given that ECG foundation models have shown that waveform representations encode risk information well beyond traditional diagnostic categories~\citep{raghunath2020prediction,li2024ecgfounder,tsai2023ecg}.

Finally, a language model without an explicit retrieval mechanism cannot consult representations of similar patients in the target population. Such representations may provide useful context beyond the current encounter. Retrieval-augmented generation has proven effective for grounding language models in external text corpora~\citep{lewis2020retrieval} and has been applied to clinical NLP with knowledge graphs~\citep{zhu2024emerge}. We study retrieval from structured patient populations, where the retrieved evidence consists of learned encounter representations.

We introduce \textbf{PRISM} (\textbf{P}opulation-\textbf{R}etrieved \textbf{I}nference for \textbf{S}tructured \textbf{M}ultimodal Prognosis)\footnote{Code: \url{https://github.com/hamzashafiq28/PRISM}.}, a multimodal architecture that supplies a language model with population-level context while representing structured clinical measurements explicitly. Rather than treating the language model as a self-contained reasoner that must absorb all clinical context through its tokenizer, PRISM equips it with three external capacities: a dense numerical encoder that represents every clinical feature as an individual learned token, a temporal aggregation mechanism that distills longitudinal waveform history into a compact representation under exponential recency weighting, and a population retrieval module that supplies representations of historically similar patients as non-parametric context. Each capacity is realized through a dedicated component: an FT-Transformer~\citep{gorishniy2021revisiting} for structured EHR encoding, an exponential recency-weighting module for ECG history aggregation, and a dual-modality population memory bank over ECG and EHR embeddings. Their outputs are projected as soft tokens into a shared sequence consumed by a medical language model (i.e., MedGemma~\citep{sellergren2025medgemma}) fine-tuned with low-rank adaptation~\citep{hu2022lora}, and a single trained model answers all 1{,}443 task questions,
each in its own forward pass, without task-specific output heads.
The mechanisms we build on are established individually, and
per-feature numerical encoding in particular is standard practice in
tabular and relational deep learning~\citep{gorishniy2021revisiting}.
What we contribute is their composition into a single tokenized
interface over heterogeneous clinical evidence, together with an
empirical account of how performance changes as individual evidence
streams are removed or replaced across task groups.

We evaluate PRISM on the MDS-ED benchmark~\citep{lopezalcaraz2025mdsed}, the most comprehensive multimodal clinical prognosis benchmark available, comprising 1{,}443 binary prediction tasks over raw 12-lead ECG waveforms and 470 structured clinical variables from MIMIC-IV~\citep{johnson2023mimic}. PRISM attains the strongest reported figures across all four task groups: 86.24\% on Diagnosis (1{,}428 ICD-10 tasks), 93.12\% on Deterioration (6~tasks), 91.79\% on ICU admission (2~tasks), and 92.82\% on Mortality (7~tasks). Counting tasks whose lower 95\% confidence bound exceeds an AUROC of 0.80, the criterion used by UniPACT, PRISM qualifies on 1{,}145 of 1{,}443 tasks (79.3\%) against 883 for UniPACT (61.2\%) and 623 for MDS-ED (43.2\%), gains of 262 and 522 tasks respectively. Controlled ablations quantify the corresponding changes: full text serialization in place of the numerical EHR pathway is $-3.66$ AUROC points, removing the pre-arrival window $-1.97$, removing the current ECG $-1.61$, removing longitudinal history $-1.22$, and zeroing retrieval $-0.99$. We additionally evaluate external transfer and supervised adaptation on MC-MED~\citep{PhysioNet-mc-med-1.0.1}, with controlled locally trained baselines. Fine-tuned PRISM has the highest overall point estimate in this external comparison, and supplementary temporal validation compares PRISM with compact fusion baselines and retrieval controls. Our principal contributions are:
\begin{itemize}[leftmargin=1.5em,itemsep=2pt,topsep=4pt]
    \item We propose PRISM, a multimodal architecture for clinical prognosis that unifies dense numerical encoding of structured clinical records, temporal aggregation of longitudinal waveform history, and population-level retrieval-augmented generation within a single language model reasoning framework. Against a 5.9M-parameter fusion model receiving the same five input streams and frozen encoders, PRISM improves AUROC by 14.40 points, and by 13.33 points against the strongest compact three-stream configuration. These gains compare complete model configurations under the controls described in Section~\ref{sec:nollm}.
    \item We show that replacing the compact numerical soft-token pathway with full text serialization of the same structured record reduces AUROC by 3.66 points, with the largest differences on diagnosis tasks that depend strongly on continuous-valued laboratory measurements.
    \item We introduce a temporal history aggregator that pools prior-visit ECG embeddings under exponential recency weighting. On the 122 encounters where pooling is operative, the learned rule has the highest point estimate among the tested alternatives: +0.67 AUROC points over the best fixed decay, +1.10 over uniform averaging, and up to +2.33 over single-acquisition selection.
    \item We extend retrieval-augmented generation from text corpora to empirical patient populations by indexing per-encounter ECG and EHR embeddings and injecting mean-pooled neighbor representations as additional soft-token context. Retrieval operates entirely in representation space, so the model gains population context while no individual retrieved record enters the prompt.
\end{itemize}
\begin{figure}[t]
    \centering
    \captionsetup{font=scriptsize}
    \includegraphics[width=\textwidth]{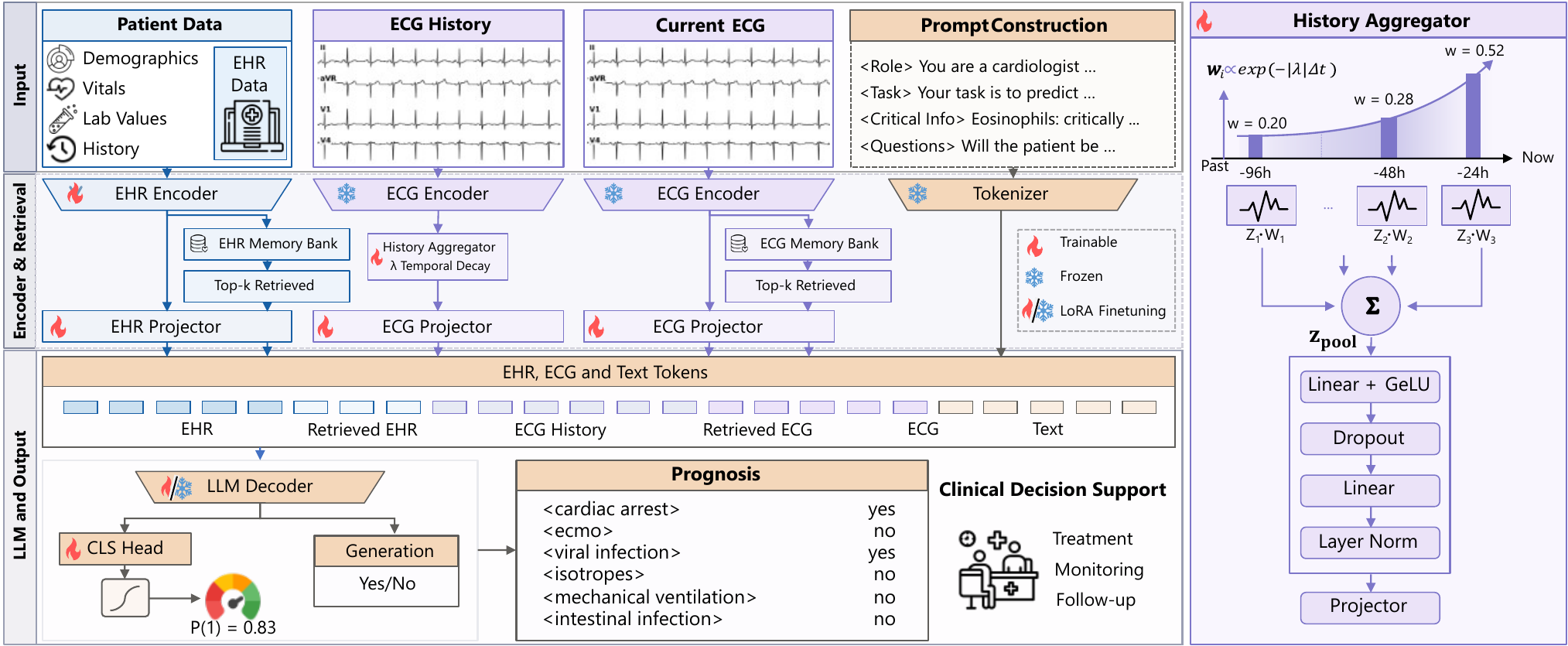}
    \caption{Overview of the PRISM architecture. Five modality streams (structured EHR, current ECG, longitudinal ECG history, retrieved EHR context, and retrieved ECG context) are independently encoded and projected as soft tokens into a shared sequence consumed by a LoRA-adapted MedGemma-4B decoder. The population memory bank (center) retrieves the $k{=}3$ nearest training-fold encounters from dual EHR and ECG indices constructed after Stage~1 and passes their mean-pooled embeddings, not their outcome labels, to the decoder. The temporal history aggregator (right) pools prior ECG embeddings under exponential recency weighting. Each task question is answered in one forward pass, via either the generative or the classification pathway, with no task-specific heads.}
    \label{fig:architecture}
\end{figure}
\section{Related Work}
\paragraph{Multimodal clinical LLMs and structured data encoding.}
The soft-token alignment paradigm, in which encoder outputs are projected into a language model's embedding space as learnable tokens~\citep{liu2023llava,li2023llavamed}, has become the dominant approach for multimodal clinical prediction. Recent systems apply this paradigm to align ECG waveform encoders~\citep{li2024ecgfounder,tian2024ked,hung2025boosting} with LLM backbones for diagnostic and prognostic tasks~\citep{tang2026unipact,pham2025qheart,zhao2024ecgchat}. A persistent design choice across these systems is that structured clinical data, including laboratory values, vital signs, and demographics, enters the language model as serialized natural-language strings rather than through encoders that preserve numerical fidelity. Transformer architectures designed for heterogeneous tabular data, such as the FT-Transformer~\citep{gorishniy2021revisiting}, map each feature to an individual token via learned per-feature projections. This provides a numerical encoding mechanism for multimodal clinical prediction. Separately, pre-trained EHR Transformers such as Med-BERT~\citep{rasmy2021medbert} and BEHRT~\citep{li2020behrt} learn effective representations from longitudinal diagnostic code sequences but target discrete event vocabularies rather than the dense continuous-valued features that characterize acute-care data. We integrate an FT-Transformer based encoder directly into the multimodal LLM pipeline, encoding all 470 clinical features numerically and providing an explicit numerical representation alongside the textual prompt.
\paragraph{Longitudinal modeling in clinical prediction.}
Physiological signals acquired serially carry prognostic information beyond any single measurement. In the ECG domain, serial waveform evolution provides independent predictive value for adverse cardiac events in acute presentations~\citep{lehmacher2020serial}, and large-scale studies have confirmed that ECG waveforms encode risk patterns invisible to expert interpretation~\citep{raghunath2020prediction,tsai2023ecg}. Existing multimodal clinical LLMs, however, encode only the index-visit acquisition, treating each encounter in isolation. Longitudinal EHR models~\citep{rasmy2021medbert,li2020behrt} capture temporal dependencies over months or years of discrete clinical events, but the problem of incorporating short-horizon, multi-acquisition waveform histories within a language model architecture remains unaddressed. We address this gap through a temporal history aggregator that pools prior-visit ECG embeddings via exponential recency weighting governed by a single scalar, injecting longitudinal waveform context into the language model at negligible parameter cost.
\paragraph{Retrieval-augmented reasoning in healthcare.}
Retrieval-augmented generation (or RAG) improves factual consistency by grounding language model outputs in external evidence~\citep{lewis2020retrieval}. In clinical settings, this motivation is particularly strong: even large language models encoding substantial medical knowledge exhibit factual inconsistencies under expert scrutiny~\citep{singhal2023large}. EMERGE~\citep{zhu2024emerge} applies retrieval over a clinical knowledge graph to generate task-relevant patient summaries for mortality and readmission prediction. Clinical RAG commonly retrieves from knowledge sources such as guidelines, ontologies, or literature. Patient-population retrieval instead supplies per-encounter embeddings as context. We investigate this setting by indexing learned encounter representations from the training cohort in a dual-modality memory bank and supplying mean-pooled representations of clinically similar cases as additional context for the language model.
\section{Method}
\label{sec:method}
\subsection{Problem Formulation and Overview}
\label{sec:overview}
Let each clinical encounter be characterized by a tuple $(\mathbf{X}_{\mathrm{ehr}},\, \mathbf{X}_{\mathrm{ecg}},\, \mathcal{H},\, \mathbf{p})$, where $\mathbf{X}_{\mathrm{ehr}} \in \mathbb{R}^{D}$ is a $D$-dimensional structured clinical feature vector (here $D{=}470$, comprising laboratory analytes, vital signs, demographics, and categorical indicators), $\mathbf{X}_{\mathrm{ecg}} \in \mathbb{R}^{12 \times T}$ is a raw 12-lead ECG waveform of length $T$, $\mathcal{H} = \{(\mathbf{X}_{\mathrm{ecg}}^{(i)}, \delta_i)\}_{i=1}^{N}$ is a set of $N \geq 0$ prior ECG acquisitions with corresponding temporal offsets $\delta_i > 0$ (in hours before the index visit), and $\mathbf{p}$ is a natural-language clinical prompt specifying the prediction target. Given a set of $\mathcal{T}$ binary outcome tasks (here $|\mathcal{T}|{=}1{,}443$), the goal is to learn a function $f$ that maps each encounter to a vector of probabilities $\hat{\mathbf{y}} = f(\mathbf{X}_{\mathrm{ehr}}, \mathbf{X}_{\mathrm{ecg}}, \mathcal{H}, \mathbf{p}) \in [0,1]^{|\mathcal{T}|}$, where each $\hat{y}_t$ estimates $P(y_t {=} 1)$ for outcome $t$. PRISM decomposes this function into five encoding streams, a population retrieval mechanism, and a shared language model decoder, as illustrated in Figure~\ref{fig:architecture}. Each stream produces a fixed-length embedding that is projected as a sequence of soft tokens into the language model's embedding space. Prompt design and data construction are described in Appendices \ref{app:prompt} and \ref{app:data_construction}.
\subsection{Structured EHR Encoding}
\label{sec:ehr_enc}
We encode the complete $D$-dimensional clinical feature vector using a Feature Tokenizer Transformer (FT-Transformer)~\citep{gorishniy2021revisiting}, which preserves the numerical fidelity of each clinical measurement by mapping it to an individual learned token rather than serializing it as text.
Concretely, each continuous feature $x_j$ is projected to a $d_{\mathrm{model}}$-dimensional token via a learned per-feature affine transformation $\mathbf{t}_j = x_j \mathbf{w}_j + \mathbf{b}_j$, where $\mathbf{w}_j \in \mathbb{R}^{d_{\mathrm{model}}}$ and $\mathbf{b}_j \in \mathbb{R}^{d_{\mathrm{model}}}$ are feature-specific parameters. Categorical features are embedded through trainable lookup tables. A learnable \texttt{[CLS]} token $\mathbf{t}_{\mathrm{cls}}$ is prepended to the resulting sequence $[\mathbf{t}_{\mathrm{cls}},\, \mathbf{t}_1,\, \dots,\, \mathbf{t}_D]$, which is processed by an $L$-layer Transformer encoder with $d_{\mathrm{model}}{=}192$, $H{=}4$ attention heads, and pre-layer normalization~\citep{xiong2020prelayernorm}. The \texttt{[CLS]} output is projected through a two-layer feed-forward network (FFN) to produce the EHR embedding:
\begin{equation}
  \mathbf{z}_{\mathrm{EHR}} = \mathrm{FFN}\!\left(\mathrm{Transformer}^{L}\!\left([\mathbf{t}_{\mathrm{cls}},\, \mathbf{t}_1, \dots, \mathbf{t}_D]\right)_{\mathrm{cls}}\right) \;\in\; \mathbb{R}^{512}.
  \label{eq:ehr}
\end{equation}
To capture short-horizon physiological trends beyond the index measurement, 96-hour prior vital sign histories are summarized per channel into seven aggregate statistics (minimum, maximum, mean, standard deviation, first, last, and slope), yielding 42 supplementary features concatenated to the base vector before encoding. Missing values are imputed with training-fold medians to prevent information leakage across data splits. The EHR encoder is trained end-to-end during Stage~1 and frozen thereafter.
\paragraph{Input windows.}
Two windows govern the inputs to PRISM, and both are defined relative
to ED arrival, denoted $t_0$. The MDS-ED benchmark extracts its
feature vector from the interval $[t_0,\, t_0 + 90\,\mathrm{min}]$,
and this interval also fixes the prediction cutoff. No information
recorded after $t_0 + 90\,\mathrm{min}$ enters PRISM through any
stream, so the model sees nothing that postdates the point of
prediction. The supplementary vital sign statistics described above
and the prior ECG acquisitions described in Section~\ref{sec:history}
are drawn instead from the retrospective interval
$[t_0 - 96\,\mathrm{h},\, t_0]$, which lies wholly before arrival.
This design extends the evidence available at the point of prediction
without crossing the cutoff. Prior systems evaluated on MDS-ED use
only the 90-minute post-arrival window, so the pre-arrival record is
signal that those systems leave unused. We measure its contribution
directly by retraining PRISM under the benchmark window alone, and
Section~\ref{sec:benchmark} reports the resulting matched-window comparison.
\subsection{ECG Waveform Encoding}
\label{sec:ecg_enc}
The index-visit ECG waveform $\mathbf{X}_{\mathrm{ecg}} \in \mathbb{R}^{12 \times T}$ is encoded by D-BETA~\citep{hung2025boosting}, a waveform-language foundation model pre-trained through two complementary objectives: masked modality modeling, which reconstructs masked segments of both the waveform and paired clinical text, and a contrastive objective with a nearest-neighbor negative sampling strategy that aligns waveform and text representations in a shared embedding space. The encoder produces a fixed-dimensional waveform embedding $\mathbf{z}_{\mathrm{ECG}} \in \mathbb{R}^{d_{\mathrm{ecg}}}$.
D-BETA is selected as default ECG encoder for its joint waveform-text pre-training, which organizes the embedding space by clinical co-occurrence. This geometric property benefits both patient-level encoding and population memory bank retrieval (see Section~\ref{sec:memory_bank}). The encoder remains frozen throughout both training stages, with embeddings pre-computed offline.
\subsection{Temporal History Aggregation}
\label{sec:history}
Serial ECG changes carry prognostic information beyond any single acquisition~\citep{lehmacher2020serial}. To incorporate this longitudinal context, we introduce a temporal history aggregator that weights prior-visit ECG embeddings by their recency relative to the index encounter (Figure~\ref{fig:architecture}, right panel).
For a patient with $N$ prior acquisitions obtained $\delta_i$ hours before the index visit, the aggregator computes normalized exponential recency-decay weights:
\begin{equation}
  w_i = \frac{\exp\!\bigl(-|\lambda|\;\delta_i\,/\,\tau\bigr)}
             {\sum_{j=1}^{N}\exp\!\bigl(-|\lambda|\;\delta_j\,/\,\tau\bigr)},
  \label{eq:decay}
\end{equation}
where $\lambda \in \mathbb{R}$ is a single learnable scalar initialized to $1.0$ and $\tau = 96$ hours is a fixed normalization constant corresponding to the history window. The absolute value $|\lambda|$ enforces non-negative decay while permitting unconstrained gradient flow. The pooled history representation is then:
\begin{equation}
  \mathbf{z}_{\mathrm{pool}} = \sum_{i=1}^{N} w_i\;\mathbf{z}_{\mathrm{ECG}}^{(i)} \;\in\; \mathbb{R}^{d_{\mathrm{ecg}}}.
  \label{eq:pool}
\end{equation}
This weighted average is transformed by an adapter network comprising a linear projection, GELU activation, dropout ($p{=}0.1$), a second linear projection to 512 dimensions, and layer normalization:
\begin{equation}
  \mathbf{z}_{\mathrm{hist}} = \mathrm{LayerNorm}\!\left(\mathbf{W}_2\;\mathrm{Dropout}\!\left(\mathrm{GELU}\!\left(\mathbf{W}_1\;\mathbf{z}_{\mathrm{pool}} + \mathbf{b}_1\right)\right) + \mathbf{b}_2\right) \;\in\; \mathbb{R}^{512}.
  \label{eq:hist_adapter}
\end{equation}
For patients without prior acquisitions ($N{=}0$), a separate learnable no-history embedding initialized from $\mathcal{N}(0, 0.02^2)$ replaces $\mathbf{z}_{\mathrm{hist}}$, allowing the model to distinguish the absence of history from a genuine temporal representation. The entire aggregation module, including $\lambda$, is trained during Stage~1 and frozen in Stage~2.
Training moves $|\lambda|$ from its initialization of $1.0$ to
$0.9891$, which corresponds to a half-life of 67.3 hours over the
96-hour window. The resulting weighting is graded rather than
saturating. Two acquisitions separated by one hour receive weights in
the ratio $1.01{:}1$, at twelve hours $1.13{:}1$, at two days
$1.64{:}1$, and across the full window $2.69{:}1$. The aggregator
therefore behaves close to a uniform average for acquisitions taken
minutes apart and discriminates increasingly as the separation grows
to hours and days, reflecting the recency prior when acquisition times are irregular.
The scope of the mechanism is set by acquisition availability. For the
89.3\% of encounters with no prior acquisition the learned no-history
embedding is used, and for the 8.8\% with exactly one the normalized
weight is $1$ irrespective of $\lambda$. The pooling rule is therefore
operative for the 1.9\% of encounters carrying two or more
acquisitions, and Section~\ref{sec:ablation} evaluates it on exactly
that population. Appendix~\ref{app:decay} reports the availability
distribution and the derived decay quantities.
\subsection{Population Memory Bank}
\label{sec:memory_bank}
To provide population-level retrieval context, PRISM constructs a dual-modality memory bank using FAISS \citep{douze2024faiss} over per-encounter embeddings from the training cohort. After Stage~1 training, the learned EHR and ECG encoders produce fixed-length representations for every training encounter. These embeddings are $\ell_2$-normalized and stored in two independent nearest-neighbor indices: an EHR index operating over $\mathbf{z}_{\mathrm{EHR}} \in \mathbb{R}^{512}$ for EHR-derived similarity, and an ECG index operating over $\mathbf{z}_{\mathrm{ECG}} \in \mathbb{R}^{d_{\mathrm{ecg}}}$ for ECG-derived similarity.
At inference, the query patient's EHR and ECG embeddings are used to retrieve the $k$ nearest neighbors independently from each index using cosine similarity. The top-$k$ retrieved embeddings from each modality are aggregated via mean pooling to produce context vectors:
\begin{equation}
  \mathbf{c}_{\mathrm{EHR}} = \frac{1}{k}\sum_{i=1}^{k} \mathbf{z}_{\mathrm{EHR}}^{(\mathrm{nn}_i)} \;\in\; \mathbb{R}^{512}, \qquad
  \mathbf{c}_{\mathrm{ECG}} = \frac{1}{k}\sum_{i=1}^{k} \mathbf{z}_{\mathrm{ECG}}^{(\mathrm{nn}_i)} \;\in\; \mathbb{R}^{d_{\mathrm{ecg}}},
  \label{eq:retrieval}
\end{equation}
where $\mathrm{nn}_i$ denotes the $i$-th nearest neighbor. The neighborhood size $k{=}3$ is selected via held-out sensitivity analysis over $k \in \{1, 3, 5\}$ and held fixed across all task groups. To eliminate retrieval latency during training and inference, neighbor assignments are materialized as pre-computed row-aligned lookup tables after Stage~1, reducing each retrieval operation to a single array index. Appendix~\ref{app:memory_bank} provides a complete configuration of the retrieval index and its key details.
\paragraph{Retrieved content.}
Equation~\ref{eq:retrieval} specifies the retrieved context
completely. The model receives two vectors per encounter,
$\mathbf{c}_{\mathrm{EHR}}$ and $\mathbf{c}_{\mathrm{ECG}}$, and each
is the unweighted mean of $k{=}3$ neighbor embeddings drawn from its
own modality-specific index. Retrieval therefore operates entirely in
representation space. Stage~1 training on the prediction objective
shapes the geometry of these embeddings, so outcome-related
structure may be reflected indirectly in the pooled vectors; however,
no label, code, feature value, identifier or descriptive text from a
neighbor enters the prompt. The two indices are queried
independently and pass through separate projectors into separate
soft-token positions, so the decoder receives EHR-derived and
ECG-derived population context as distinct inputs.
\paragraph{Same-patient retrieval.}
The MDS-ED partitions are disjoint at the level of the patient rather
than the encounter, and the memory bank is built only from
training-fold encounters. A validation or test query therefore cannot
return an encounter belonging to the same patient, because no
encounter of that patient is present in the index, and the
same-patient retrieval rate at evaluation is exactly zero.
Within the training fold the situation is different in principle,
since a patient with several encounters may retrieve one of their own,
and we quantified this directly over all training and validation queries.
At the deployed neighborhood size of $k{=}3$, 29 queries (0.0237\%)
retrieve an encounter belonging to the same subject, against roughly
0.01\% expected under random retrieval. This remains a very small
fraction despite 76{,}875 rows in the combined training and validation cohort (62.7\% of 122{,}641 encounters) belonging
to patients with multiple encounters; the cause of this small
enrichment is not established. A related qualification concerns the
exclusion mechanism. Self-matches
are removed by dropping the rank-0 neighbor, which assumes the query
row always ranks first against itself. In 0.0073\% of queries it does
not, because a near-duplicate embedding displaces it. Filtering on
encounter identifier would remove this residual entirely. Accordingly,
the implemented procedure is described as rank-0 removal rather than
guaranteed self-exclusion. Because evaluation is patient-disjoint, the
same-subject retrieval rate is 0 at evaluation; the small training-fold
rate makes same-patient lookup unlikely to explain the observed test
benefit.
The dual-index design keeps EHR- and ECG-derived retrieval context separate. As the ablation study in Section~\ref{sec:ablation} shows, neither index alone materially improves over no retrieval, whereas supplying both together does. These controls establish an interaction in the deployed PRISM configuration but do not identify its mechanism; an explicit joint-similarity retrieval comparison remains future work.
\subsection{Multimodal Fusion and Language Model Adaptation}
\label{sec:fusion}
\paragraph{Soft-token projection.}
Each modality embedding is mapped to a sequence of $n_m$ soft tokens in the language model's embedding space through a dedicated two-layer projector:
\begin{equation}
  \mathbf{S}_m = \mathrm{Reshape}_{n_m \times d_{\mathrm{llm}}}\!\left(
    \mathbf{W}_m^{(2)}\;\mathrm{GELU}\!\left(
      \mathbf{W}_m^{(1)}\;\mathrm{LayerNorm}(\mathbf{z}_m) + \mathbf{b}_m^{(1)}
    \right) + \mathbf{b}_m^{(2)}
  \right),
  \label{eq:softtok}
\end{equation}
where $\mathbf{W}_m^{(1)} \in \mathbb{R}^{1024 \times d_{\mathrm{in}}}$, $\mathbf{W}_m^{(2)} \in \mathbb{R}^{n_m d_{\mathrm{llm}} \times 1024}$, $n_m{=}8$ for all five modality streams, and $d_{\mathrm{llm}}{=}2560$ for MedGemma-4B. The five modality streams are: current ECG ($\mathbf{z}_{\mathrm{ECG}}$), structured EHR ($\mathbf{z}_{\mathrm{EHR}}$), longitudinal ECG history ($\mathbf{z}_{\mathrm{hist}}$), retrieved EHR context ($\mathbf{c}_{\mathrm{EHR}}$), and retrieved ECG context ($\mathbf{c}_{\mathrm{ECG}}$). Appendix~\ref{app:proj} tabulates complete per-modality input and output dimensions for all five projectors.
\paragraph{Placeholder substitution.}
The tokenized clinical prompt $\mathbf{p}$ contains four or five designated placeholder tokens that serve as positional anchors within the text sequence, the fifth present only when the encounter has at least one prior ECG acquisition. At forward time, each placeholder is replaced in-place by its corresponding $n_m \times d_{\mathrm{llm}}$ soft-token matrix $\mathbf{S}_m$, while all remaining positions are embedded through the standard LLM vocabulary layer. The resulting sequence interleaves 32 or 40 modality tokens ($4 \times 8$ or $5 \times 8$, according to whether prior ECG history is available) with a variable number of text tokens into a unified embedding tensor $\mathbf{E} \in \mathbb{R}^{S \times d_{\mathrm{llm}}}$ consumed directly by the decoder. This substitution-based alignment enables cross-modal reasoning within the LLM's self-attention layers without architectural modification.
\paragraph{Language model backbone and adaptation.}
We use MedGemma-4B~\citep{sellergren2025medgemma}, a decoder-based four-billion-parameter language model pre-trained on de-identified biomedical corpora. During Stage~1, all LLM parameters are frozen and only the encoders and projectors are optimized. During Stage~2, we apply Low-Rank Adaptation (LoRA)~\citep{hu2022lora} to the query, key, value, output, gate, up-projection, and down-projection matrices of every self-attention layer with rank $r{=}16$, scaling factor $\alpha_{\mathrm{LoRA}}{=}32$, and dropout $0.05$.
\subsection{Training and Inference}
\label{sec:training}
PRISM is trained in two stages, each with a distinct set of trainable parameters and a shared composite objective.
\paragraph{Composite objective.}
Both stages minimize a weighted combination of causal language modeling loss on the answer tokens and a binary classification auxiliary loss:
\begin{equation}
  \mathcal{L} = \mathcal{L}_{\mathrm{CLM}} + \alpha\;\mathcal{L}_{\mathrm{BCE}},
  \label{eq:loss}
\end{equation}
where $\mathcal{L}_{\mathrm{CLM}}$ is the standard next-token cross-entropy restricted to the answer span, $\mathcal{L}_{\mathrm{BCE}}$ is binary cross-entropy applied through a linear classification head over the mean-pooled decoder hidden states at all prompt positions, and $\alpha{=}0.1$ throughout.
\paragraph{Stage~1: encoder and projector training.}
The EHR FT-Transformer, the temporal history aggregator (including $\lambda$), and the three primary projectors (ECG, EHR, history) are jointly optimized while the D-BETA encoder and MedGemma-4B backbone remain frozen. Training uses AdamW optimizer with learning rate $2 {\times} 10^{-4}$, cosine decay with 100 warm-up steps, gradient clipping at $\|\mathbf{g}\|_2 \leq 1.0$, effective batch size 32 per device (micro-batch 4, gradient accumulation over 8 steps), and BF16 mixed precision on two NVIDIA H100 80\,GB GPUs. Upon completion, the population memory bank indices are constructed from the trained encoder outputs.
\paragraph{Stage~2: retrieval-augmented language model adaptation.}
All Stage~1 parameters are frozen. The LoRA adapters, the two retrieval projectors (initialized from scratch), and the classification head (loaded from the Stage~1 checkpoint) are jointly optimized. The learning rate is reduced to $2 {\times} 10^{-5}$; the micro-batch size is reduced to two to accommodate the expanded five-modality token sequence within the 2{,}048-token budget. Checkpoints are selected by minimum validation loss on the held-out split. Full hyperparameter specifications are provided in Appendix~\ref{app:hyperparams}.
\paragraph{Inference pathways.}
At inference, PRISM supports two complementary output pathways evaluated on every task. The \emph{generative pathway} applies a softmax to the Yes/No answer-token logits at the final prompt position and uses the Yes probability as the prediction score, reported as $\text{PRISM}_{\text{gen}}$. The \emph{discriminative pathway} applies the linear head to the mean-pooled decoder hidden states, producing a continuous probability estimate $\hat{y} = P(y{=}1) \in [0,1]$ without autoregressive decoding, reported as $\text{PRISM}_{\text{disc}}$. Both pathways are reported for the primary benchmark; the external and temporal comparisons report the generative pathway.
\begin{table*}[t]
\centering
\captionsetup{font=scriptsize}
\caption{\textbf{Benchmark results across all 1{,}443 MDS-ED tasks.} AUROC (\%) is macro-averaged within each task group: Diagnosis (1{,}428 ICD-10 discharge tasks), Deterioration (6 tasks), ICU admission (2 tasks), and Mortality (7 tasks). PRISM has the highest reported values among the listed methods in each task group. Its full-input configuration additionally uses 96-hour pre-arrival history; the matched-window comparison is reported in the text. \underline{\textbf{Bold with underline}}: best. $^\dagger$Zero-shot, text-only input.}
\scriptsize
\resizebox{\textwidth}{!}{%
\begin{tabular}{@{} l S[table-format=2.2] S[table-format=2.2] S[table-format=2.2] S[table-format=2.2] @{}}
\toprule
\multirow{2}{*}{\textbf{Method}} & \multicolumn{4}{c}{\textbf{Tasks · AUROC (\%)}} \\
\cmidrule(lr){2-5}
& {\textbf{Diagnosis}} & {\textbf{Deterioration}} & {\textbf{ICU}} & {\textbf{Mortality}} \\
\midrule
\rowcolor{catgray}
\multicolumn{5}{@{}l}{\textbf{\quad ECG Foundation Models} {\tiny\color{gray}\textit{· waveform input}}} \\
\cmidrule(lr){1-5}
\quad HeartBERT~\citep{tahery2024heartbert} & 56.35 & 58.24 & 57.28 & 61.11 \\
\quad ECGFM-KED~\citep{tian2024ked} & 58.75 & 64.76 & 61.32 & 65.76 \\
\quad ESI~\citep{yu2024esi} & 60.98 & 63.34 & 62.52 & 66.28 \\
\quad ECGFounder~\citep{li2024ecgfounder} & 61.52 & 69.35 & 64.13 & 69.42 \\
\addlinespace[0.5em]
\rowcolor{catgray}
\multicolumn{5}{@{}l}{\textbf{\quad Image-Based Models} {\tiny\color{gray}\textit{· rendered ECG input}}} \\
\cmidrule(lr){1-5}
\quad GEM~\citep{lan2025gem} & 61.52 & 55.80 & 60.40 & 65.70 \\
\quad PULSE~\citep{liu2024pulse} & 58.47 & 61.10 & 64.70 & 68.70 \\
\quad TimeMaster~\citep{zhang2025timemaster} & 57.91 & 60.80 & 63.80 & 57.30 \\
\quad ECG-Instruct~\citep{nandakishor2025highaccuracy} & 47.33 & 46.00 & 44.40 & 43.70 \\
\addlinespace[0.5em]
\rowcolor{catgray}
\multicolumn{5}{@{}l}{\textbf{\quad Signal-Based QA Models} {\tiny\color{gray}\textit{· ECG waveform input}}} \\
\cmidrule(lr){1-5}
\quad ECGQA-FSL~\citep{tang2024ecgqafsl} & 50.56 & 50.46 & 51.06 & 52.15 \\
\quad Q-HEART~\citep{pham2025qheart} & 50.15 & 55.42 & 54.89 & 56.32 \\
\quad ECG-Chat~\citep{zhao2024ecgchat} & 49.70 & 57.54 & 56.40 & 55.78 \\
\addlinespace[0.5em]
\rowcolor{catgray}
\multicolumn{5}{@{}l}{\textbf{\quad Frontier LLMs$^\dagger$} {\tiny\color{gray}\textit{· text-only input}}} \\
\cmidrule(lr){1-5}
\quad GPT-5$^\dagger$~\citep{openai2025gpt5} & 55.68 & 78.12 & 70.04 & 62.27 \\
\quad Gemini-2.5 Pro$^\dagger$~\citep{comanici2025gemini25pushingfrontier} & 57.82 & 78.72 & 70.70 & 72.38 \\
\addlinespace[0.5em]
\rowcolor{catgray}
\multicolumn{5}{@{}l}{\textbf{\quad Multimodal Baselines} {\tiny\color{gray}\textit{· waveform + EHR}}} \\
\cmidrule(lr){1-5}
\quad MDS-ED~\citep{lopezalcaraz2025mdsed} & 82.56 & 90.70 & 90.63 & 91.68 \\
\quad UniPACT~\citep{tang2026unipact} & 83.98 & 91.17 & 90.50 & 91.82 \\
\midrule
\rowcolor{highlightgreen}
\multicolumn{5}{@{}l}{\textbf{\quad PRISM (Ours)} {\tiny\color{gray}\textit{· waveform + EHR + history + retrieval}}} \\
\cmidrule(lr){1-5}
\rowcolor{highlightgreen!20}
\quad $\textbf{PRISM}_{\textbf{gen}}$ & \underline{\textbf{86.24}} & \textbf{92.90} & \underline{\textbf{91.79}} & \underline{\textbf{92.82}} \\
\rowcolor{highlightgreen!20}
\quad $\textbf{PRISM}_{\textbf{disc}}$ & \textbf{86.06} & \underline{\textbf{93.12}} & \textbf{91.55} & \textbf{92.79} \\
\bottomrule
\end{tabular}
}
\label{tab:benchmark}
\end{table*}
\section{Results}
\label{sec:results}
\subsection{Overall Benchmark Performance}
\label{sec:benchmark}
The primary benchmark experiments use the standardized MDS-ED splits; external evaluation is reported in Section~\ref{sec:mcmed_external}, with supplementary temporal validation in Appendix~\ref{app:temporal_generalization}. Following Stage 1 and Stage 2 optimization on 4.5 million training samples, the benchmark test partition contains 6,454 encounters. The sampled evaluation retains 262{,}158 encounter--task records after removing records from 83 encounters with unavailable ECG waveforms; exact class balance is therefore not guaranteed after exclusion. Table~\ref{tab:benchmark} reports AUROC across all four MDS-ED task groups for PRISM and 15 baseline systems spanning five model families. PRISM attains the strongest reported figures on every task group: \textbf{86.24\%} on Diagnosis (1{,}428 ICD-10 discharge tasks), \textbf{93.12\%} on Deterioration (6 tasks), \textbf{91.79\%} on ICU admission (2 tasks), and \textbf{92.82\%} on Mortality (7 tasks), with the generative and discriminative pathways achieving the best score on complementary task groups.
The supervised multimodal systems have higher diagnosis AUROC than the listed single-modality ECG models and zero-shot language models (Table~\ref{tab:benchmark}). Among the four task groups, Deterioration, ICU admission, and Mortality predict future adverse events from data available within the first 90 minutes after ED arrival, whereas Diagnosis tasks predict discharge ICD-10 labels assigned at the end of the hospital stay, a temporally forward-looking classification problem grounded in a different clinical signal.
Table~\ref{tab:benchmark} and Table~\ref{tab:pertask} report the
single-run evaluation. The three-seed evaluation used for the
control experiments yields group means of 0.8619, 0.9291, 0.9180, and
0.9279, corresponding to a macro-of-macro AUROC of 0.9092, compared
with 0.9094 in the single-run evaluation. This 0.02-point difference is
within the 0.32-point standard deviation across the three full-model
runs. Ablation effects are computed relative to the corresponding
full-model reference from the corresponding evaluation.

The two prior multimodal systems, MDS-ED~\citep{lopezalcaraz2025mdsed} and UniPACT~\citep{tang2026unipact}, fuse ECG representations with clinical context and reach 88.89\% and 89.37\% overall. Relative to UniPACT, the highest PRISM scores across the two pathways are higher by 2.26 points on Diagnosis, 1.95 on Deterioration, 1.29 on ICU admission, and 1.00 on Mortality.

PRISM uses the 96-hour pre-arrival window described in
Section~\ref{sec:ehr_enc}, whereas the reported baselines use only the
benchmark window. Removing this additional history and retraining
reduces PRISM from 0.9092 to 0.8895, a difference of 1.97 AUROC points.
Under the benchmark window alone, performance is closely spaced across
PRISM (0.8895), UniPACT (0.8937), and MDS-ED (0.8889), indicating that
the additional pre-arrival history contributes materially to PRISM's
full-input benchmark result.
Section~\ref{sec:nollm} provides a complementary matched-input
comparison against compact non-LLM fusion models receiving the same
evidence streams and frozen upstream encoders. PRISM outperforms the tested compact configurations; the comparison and its methodological scope are detailed in Section~\ref{sec:nollm}.

We estimated 95\% confidence intervals via empirical bootstrap (1{,}000 resamples per task, macro-averaged across tasks within each group). The resulting intervals are 86.24 [85.92, 86.45] for Diagnosis, 92.90 [89.16, 95.82] for Deterioration, 91.79 [90.92, 92.65] for ICU, and 92.82 [91.98, 93.50] for mortality. A complete breakdown of per-task performance for Deterioration, ICU, and Mortality, with confidence intervals, is provided in Appendix~\ref{app:pertask}.

\begingroup

\subsection{External Evaluation on MC-MED}
\label{sec:mcmed_external}
We further evaluate PRISM on MC-MED~\citep{PhysioNet-mc-med-1.0.1} from a separate institution. Table~\ref{tab:mcmed_external} distinguishes source-only transfer using the original checkpoint and memory bank, local-bank adaptation without parameter updates, and supervised local adaptation by fine-tuning PRISM. The locally trained baselines provide the primary comparison for fine-tuned PRISM; they receive target-site supervision unavailable in the source-only regime. All models are evaluated on the same eligible records and tasks within this setting.

MC-MED comprises Stanford emergency-department visits from 2020--2022. Its single measured ECG lead is repeated across the 12 encoder input channels, so this evaluation also involves a change in waveform acquisition rather than an identical diagnostic 12-lead input. Its chronological split contains 31{,}239 training, 4{,}579 validation, and 4{,}303 test visits. Evaluation uses 13{,}254 class-balanced encounter--task records from 3{,}588 patients across 306 eligible tasks (297 diagnosis, four deterioration, one ICU, and four mortality tasks). PRISM results in Table~\ref{tab:mcmed_external} use the generative pathway. Appendix~\ref{app:external_temporal_protocol} provides cohort counts, endpoint coverage, and reporting conventions.

\begin{table}[t]
\centering

\captionsetup{font=scriptsize}
\caption{\textbf{External evaluation on MC-MED.} All entries are AUROC; PRISM uses the generative pathway. Overall AUROC is the unweighted mean of the four task-group means. Brackets show reported 95\% patient-bootstrap confidence intervals. The first two rows use no target-site parameter training. The final two rows perturb retrieval at inference using the fine-tuned checkpoint and identical evaluation records. Bold identifies the highest point estimate among the supervised rows, not statistical significance.}
\label{tab:mcmed_external}
\small
\setlength{\tabcolsep}{3pt}
\renewcommand{\arraystretch}{1.25}
\resizebox{\textwidth}{!}{%
\begin{tabular}{@{}lccccc@{}}
\toprule
\textbf{Model / condition} & \textbf{Diagnosis} & \textbf{Deterioration} & \textbf{ICU} & \textbf{Mortality} & \textbf{Overall AUROC} \\
\midrule
\multicolumn{6}{@{}l}{\textit{Source-only transfer}} \\
PRISM, original bank & 0.7086 & 0.7112 & 0.8188 & 0.8977 & 0.7841 [0.751, 0.814] \\
\multicolumn{6}{@{}l}{\textit{Local-bank adaptation (no parameter updates)}} \\
PRISM, MC-MED bank & 0.7080 & 0.7133 & 0.8193 & 0.9025 & 0.7858 [0.754, 0.815] \\
\midrule
\multicolumn{6}{@{}l}{\textit{Supervised local training / adaptation}} \\
Gradient boosting & 0.7012 & 0.7509 & \textbf{0.8924} & 0.9465 & 0.8227 [0.796, 0.851] \\
Fusion MLP & 0.6184 & 0.6227 & 0.7188 & 0.8150 & 0.6937 [0.658, 0.729] \\
Fusion MLP + retrieval & 0.5847 & 0.6471 & 0.6987 & 0.7469 & 0.6694 [0.635, 0.703] \\
Fine-tuned PRISM & \textbf{0.8445} & \textbf{0.7690} & 0.8830 & \textbf{0.9479} & \textbf{0.8611} [0.834, 0.887] \\
\addlinespace[0.3em]
\multicolumn{6}{@{}l}{\textit{Inference-time controls of fine-tuned PRISM}} \\
Retrieval zeroed & 0.7920 & 0.7389 & 0.8681 & 0.9215 & 0.8301 [0.802, 0.857] \\
Random retrieval & 0.7919 & 0.7310 & 0.8657 & 0.9236 & 0.8280 [0.799, 0.856] \\
\bottomrule
\end{tabular}}
\end{table}

Replacing the original bank with an MC-MED bank changes overall AUROC from 0.7841 to 0.7858, an increase of 0.17 points. Supervised fine-tuning reaches 0.8611, an increase of 7.70 points over source-only transfer. Fine-tuned PRISM exceeds locally trained gradient boosting by 3.84 points overall, with the largest task-group advantage in diagnosis. Gradient boosting retains the higher ICU point estimate, and mortality performance is close. These results support supervised adaptation to the external setting rather than a claim that bank replacement alone resolves institutional shift. Inference-time retrieval controls provide a complementary comparison within the fine-tuned model. Zeroing retrieval lowers overall AUROC from 0.8611 to 0.8301, while random neighbors yield 0.8280, corresponding to decreases of 3.10 and 3.31 AUROC points. Genuine retrieval has higher point estimates in all four task groups. The two controls differ by only 0.21 points overall; this small difference does not establish their equivalence. The pattern is consistent with useful information in the selected neighbor representations, without identifying how the model uses that information. Group-specific intervals are reported in Appendix~\ref{app:external_temporal_protocol}, Table~\ref{tab:mcmed_retrieval_controls}.
\endgroup

\subsection{Task-Level Analysis}
\label{sec:pertask}
\begin{figure}[t]
\centering
\captionsetup{font=scriptsize}
\includegraphics[width=\textwidth]{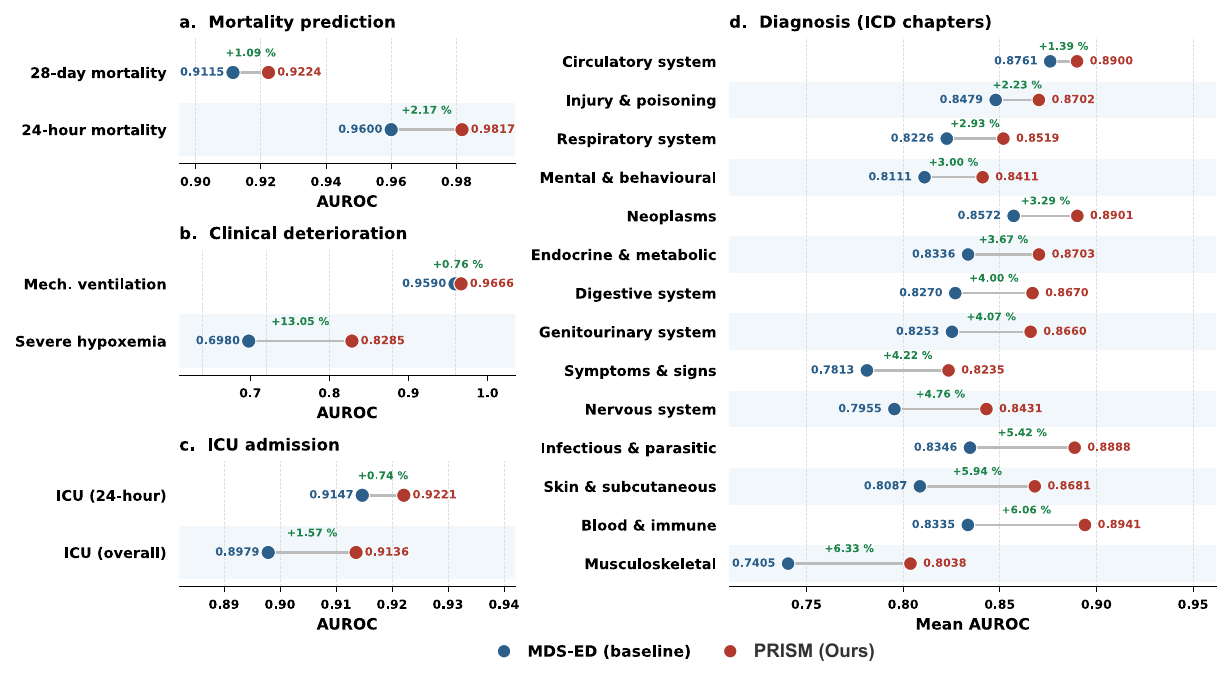}
\caption{
  \textbf{PRISM performance vs.\ MDS-ED baseline across clinical prediction categories.}
  Comparison of AUROC for (a)~mortality at multiple horizons, (b)~clinical deterioration endpoints, (c)~ICU admission, and (d)~ICD-chapter-level diagnosis groups. Red and blue markers denote PRISM and MDS-ED, respectively. Green annotations indicate relative AUROC improvement. Gains are largest on laboratory-dependent outcomes (e.g., blood and immune disorders, musculoskeletal conditions) and on deterioration endpoints where metabolic measurements are clinically relevant.
}
\label{fig:overall_results}
\end{figure}
Task-level analysis reinforces the interpretation that PRISM's gains are concentrated where structured clinical data carries the most discriminative weight. Across the full benchmark we count tasks whose lower 95\% confidence bound exceeds an AUROC of 0.80, following the criterion used by UniPACT so that the comparison is like-for-like. PRISM qualifies on 1{,}145 of 1{,}443 tasks (79.3\%), against 883 for UniPACT (61.2\%) and 623 for MDS-ED (43.2\%). This is a gain of 262 tasks over UniPACT and 522 over MDS-ED.
At the level of ICD chapters (Figure~\ref{fig:overall_results}d), improvements are consistent across all clinical domains but graded by the degree to which each domain depends on laboratory context. Circulatory system diagnoses, where ECG waveforms already carry substantial discriminative information, show a gain of 1.39\%. In contrast, musculoskeletal disorders (+6.33\%) and blood and immune system conditions (+6.06\%) show the largest improvements, consistent with these categories drawing strongly on laboratory measurements for which the numerical EHR pathway may be particularly useful.
Within the Deterioration group, the most striking individual result is on severe hypoxemia, where PRISM achieves an AUROC of 0.8285 compared to 0.6980 for MDS-ED (+13.05\%). Arterial oxygenation and respiratory measurements are clinically relevant to this endpoint and may help explain the larger observed gap. Appendix \ref{app:auroc_dist} provides a comprehensive view of how these gains distribute across the 1,428 diagnosis tasks and ICD-10 chapters.
\subsection{Population Memory Bank Analysis}
\label{sec:retrieval_analysis}
\begin{figure}[t]
    \centering
    \captionsetup{font=scriptsize}
    \includegraphics[width=\linewidth]{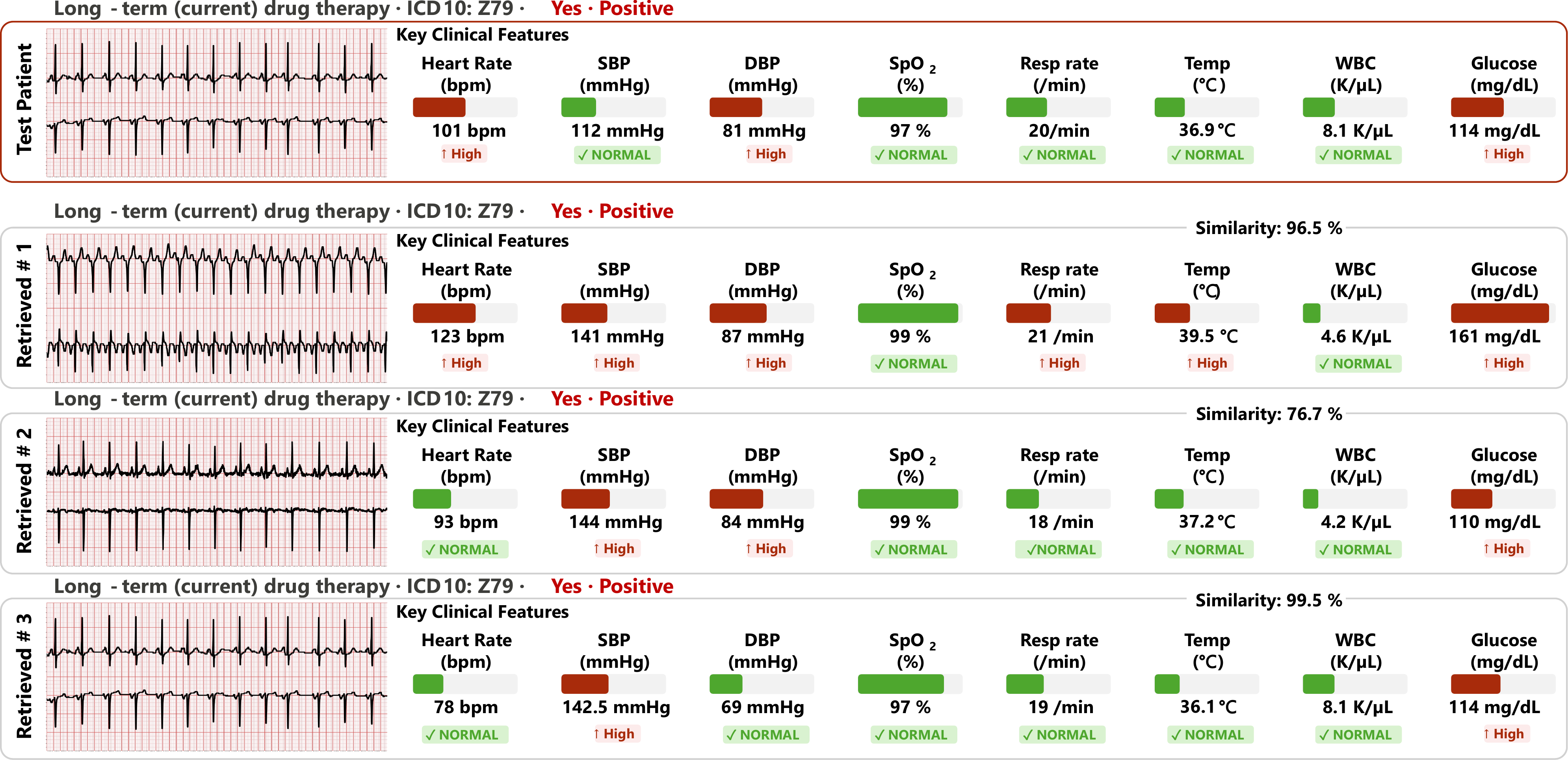}
   \caption{
    \textbf{Qualitative analysis of the population memory bank.}
    A test encounter with ICD-10 label \texttt{Z79} (long-term drug therapy, positive outcome) and three training-fold encounters selected from the dual-modality memory bank for illustration.
    The EHR and ECG indices are queried independently; the displayed encounters are returned by both indices. The model itself receives the two mean-pooled retrieval vectors defined in Equation~\ref{eq:retrieval}, rather than the individual encounters shown here.
    All displayed encounters share the same discharge label and outcome.
    Similarity values are reported within the index from which they are obtained and are not directly comparable across modalities. Shared abnormalities, including elevated glucose and heart rate, illustrate local clinical coherence in the learned embedding space.
}
    \label{fig:retrieval_case_study}
\end{figure}
Figure~\ref{fig:retrieval_case_study} presents a representative case study examining the clinical coherence of the population memory bank. For a test patient assigned ICD-10 label \texttt{Z79} (long-term drug therapy) with a positive outcome, we inspect the three training encounters whose EHR and ECG embeddings are nearest to the query in each index, with the cosine similarities shown measured within the index that returned each neighbor, since a neighbor has one similarity to the query per index and the two are not commensurable. All three retrieved neighbors share both the discharge label and the positive outcome, which illustrates local structure in the learned embedding space. The decoder conditions on the pooled vectors of Equation~\ref{eq:retrieval} rather than on these individual records, so the case is descriptive rather than evidence of cohort-level outcome concordance.
The controls in Section~\ref{sec:ablation} address the prediction
mechanism directly: genuine retrieved content improves the aggregate
result relative to random and zeroed retrieval.
\begin{table*}[t]
\centering
\captionsetup{font=scriptsize}
\caption{\textbf{Architectural and training ablations.} AUROC is
reported as macro-of-macro over the four task groups under the generative pathway. Rows marked $\dagger$ use the three-seed full-model reference ($0.9092\pm0.0032$); unmarked rows use the single-run reference (0.9094). Individual ablations are single training runs unless stated otherwise. $\boldsymbol{\Delta}$ is calculated against the corresponding full-model reference.}
\label{tab:ablation}
\small
\renewcommand{\arraystretch}{1.25}
\begin{tabular}{@{}lcc@{}}
\toprule
\textbf{Ablation / configuration} & \textbf{AUROC} & $\boldsymbol{\Delta}$ \\
\midrule
\rowcolor{catgray} \multicolumn{3}{@{}l}{\textbf{Encoder ablations}} \\
Text-serialized EHR (no tokenization)$^\dagger$ & 0.8726 & $-3.66$ \\
w/o EHR encoder & 0.8874 & $-2.20$ \\
w/o current-visit ECG$^\dagger$ & 0.8931 & $-1.61$ \\
w/o longitudinal ECG history$^\dagger$ & 0.8970 & $-1.22$ \\
\addlinespace[0.4em]
\rowcolor{catgray} \multicolumn{3}{@{}l}{\textbf{Memory bank ablations, three-seed reference}} \\
w/o retrieval context$^\dagger$ & 0.8993 & $-0.99$ \\
Retrieval with random neighbors$^\dagger$ & 0.8995 & $-0.97$ \\
\addlinespace[0.4em]
\rowcolor{catgray} \multicolumn{3}{@{}l}{\textbf{Memory bank ablations, single-run reference}} \\
w/o retrieval context & 0.8904 & $-1.90$ \\
EHR index only & 0.8910 & $-1.84$ \\
ECG index only & 0.8906 & $-1.88$ \\
\addlinespace[0.4em]
\rowcolor{catgray} \multicolumn{3}{@{}l}{\textbf{Neighborhood size}} \\
$k = 1$ & 0.8911 & $-1.83$ \\
$k = 5$ & 0.8932 & $-1.62$ \\
\addlinespace[0.4em]
\rowcolor{catgray} \multicolumn{3}{@{}l}{\textbf{Training ablations}} \\
Frozen retrieval projectors & 0.8704 & $-3.90$ \\
LoRA off (backbone frozen) & 0.8219 & $-8.75$ \\
\midrule
\rowcolor{highlightgreen}
\textbf{Full PRISM}, three-seed reference$^\dagger$ & \textbf{0.9092} & -- \\
\textbf{Full PRISM}, single-run reference & \textbf{0.9094} & -- \\
\bottomrule
\end{tabular}
\end{table*}
Inspection of key clinical features reveals plausible physiological correspondences: the test patient and all retrieved neighbors share elevated glucose and elevated heart rate as flagged abnormalities, consistent with the metabolic context associated with long-term pharmacological management in this diagnostic category. This selected case illustrates local coherence in the embedding space and does not support cohort-level inference. Because the index stores representations of real encounters, the underlying neighbors can be enumerated and reviewed offline, providing retrieval provenance without placing individual retrieved records, labels, or outcomes in the model prompt. Extended retrieval examples across four clinical phenotypes, namely metabolic disorder, chronic vascular, endocrine management and vascular risk, are provided in Appendix~\ref{app:retrieval_phenotypes}, where selected $k{=}50$ neighborhoods illustrate the index geometry. The larger neighborhood is used only to make prevalence estimates readable, and no cohort-level inference is drawn from these examples.
\subsection{Comparison with Compact Non-LLM Models}
\label{sec:nollm}
To complement the within-PRISM ablations, we train four compact
non-LLM controls using the same upstream encoders where applicable,
together with a direct $k$NN outcome-prevalence baseline. The
fusion+retrieval control receives the same five input streams as PRISM,
with the test fold and evaluation protocol held fixed.
Table~\ref{tab:nollm} reports the comparison.

\begin{table}[t]
\centering
\captionsetup{font=scriptsize}
\caption{\textbf{Architectures with and without a language model.}
The fusion+retrieval control is matched to PRISM on all five input
streams; the remaining compact controls use the subsets indicated in
their row labels. Frozen upstream encoders are shared where
applicable, and all models use the same test fold and evaluation
protocol. AUROC is reported as mean~$\pm$~standard deviation over five
seeds for the compact models and over three seeds for PRISM. Expected
calibration error uses ten equal-count bins weighted by bin size.
Average precision is reported at true cohort prevalence. Parameter
counts are totals including frozen components.}
\label{tab:nollm}
\small
\renewcommand{\arraystretch}{1.25}
\begin{tabular}{@{}lccccc@{}}
\toprule
\textbf{Model} & \textbf{LLM} & \textbf{AUROC} & \textbf{AUPRC} & \textbf{ECE} & \textbf{Params} \\
\midrule
$k$NN outcome prevalence ($k{=}50$) & No  & 0.5262 & 0.1235 & -- & 0 \\
FT-Transformer, structured EHR only & No  & $0.6870 \pm 0.0058$ & 0.1864 & 0.4362 & 3.1M \\
ECG and longitudinal history only   & No  & $0.7527 \pm 0.0031$ & 0.2378 & 0.4308 & 4.1M \\
Fusion, EHR + ECG + history         & No  & $0.7759 \pm 0.0051$ & 0.2727 & 0.4261 & 4.6M \\
Fusion + retrieval (all five streams) & No  & $0.7652 \pm 0.0043$ & 0.2500 & 0.4043 & 5.9M \\
\midrule
\rowcolor{highlightgreen}
\textbf{PRISM}                      & Yes & $\mathbf{0.9092 \pm 0.0032}$ & \textbf{0.5504} & \textbf{0.1134} & 4B \\
\bottomrule
\end{tabular}
\end{table}

The final compact control receives all five streams, including
both retrieval contexts, and is therefore matched to PRISM on input
information. PRISM exceeds this 5.9M-parameter control by 14.40 AUROC
points. Against the strongest compact configuration, the
4.6M-parameter three-stream model without retrieval, the margin is
13.33 points; average precision increases from 0.2727 to 0.5504 and
expected calibration error decreases from 0.4261 to 0.1134. These
comparisons show that the compact fusion architectures tested do not
exploit the available inputs as effectively as PRISM, but they do not
isolate the language model itself because model class, pretrained
capacity, parameter count, and optimization also differ.
Retrieval changes the compact fusion model from 0.7759 to 0.7652, and
a direct $k$NN outcome-prevalence predictor reaches 0.5262. In these
controls, retrieval does not improve the compact fusion architecture,
and the direct $k$NN rule performs near chance; both observations are
specific to the tested configurations.

The compact controls also show that both structured EHR and ECG
history carry useful signal: the EHR-only model reaches 0.6870, the
ECG+history model 0.7527, and their compact fusion 0.7759. PRISM uses a
shared-attention soft-token interface over five heterogeneous streams
while answering 1{,}443 task questions without task-specific heads,
and Table~\ref{tab:nollm} therefore compares complete model
configurations rather than isolating that design choice. The trade-off
is substantial: PRISM has roughly three orders of magnitude more
parameters than the compact fusion controls.
Parameter counts are reported for all configurations as a model-size
proxy. Cross-architecture wall-clock time is not used as a standardized
compute metric because hardware utilization, batching, and implementation
differ. For the directly comparable text-serialization experiment on
identical hardware, inference increases from roughly four hours to
twenty-four (Section~\ref{sec:ablation}).

\subsection{Ablation Study}
\label{sec:ablation}
Table~\ref{tab:streams} reports the contribution of individual
evidence streams, and Table~\ref{tab:ablation} reports the remaining
architectural and training ablations. Retraining PRISM three times
gives a standard deviation of 0.0032 AUROC. Because each ablation is a
single training run, this standard deviation is used only as a scale
reference for the observed differences rather than as a formal
standard error or significance test.

\begin{table}[t]
\centering
\captionsetup{font=scriptsize}
\caption{\textbf{Contribution of each evidence stream.} One evidence stream is removed or replaced per row, with the remaining model components held fixed. $\Delta$ is the change in AUROC
relative to full PRISM, in points. Random-neighbor and zeroed-token
controls retain the retrieval architecture while varying the content
supplied through it.}
\label{tab:streams}
\small
\renewcommand{\arraystretch}{1.25}
\begin{tabular}{@{}lcccc@{}}
\toprule
\textbf{Configuration} & \textbf{AUROC} & \textbf{AUPRC} & \textbf{ECE} & $\boldsymbol{\Delta}$ \\
\midrule
Structured EHR as serialized text & 0.8726 & 0.4632 & 0.1586 & $-3.66$ \\
w/o 96-hour pre-arrival window    & 0.8895 & 0.5358 & 0.1290 & $-1.97$ \\
w/o current-visit ECG             & 0.8931 & 0.5031 & 0.1266 & $-1.61$ \\
w/o longitudinal ECG history      & 0.8970 & 0.5418 & 0.1260 & $-1.22$ \\
w/o retrieval (tokens zeroed)     & 0.8993 & 0.5450 & 0.1239 & $-0.99$ \\
Retrieval with random neighbors  & 0.8995 & 0.5470 & 0.1224 & $-0.97$ \\
\midrule
\rowcolor{highlightgreen}
\textbf{PRISM} ($\pm 0.0032$, three seeds) & \textbf{0.9092} & \textbf{0.5504} & \textbf{0.1134} & -- \\
\bottomrule
\end{tabular}
\end{table}

Replacing the numerical EHR pathway with
full text serialization is the largest tested substitution at 3.66
points. For retrieval, zeroed tokens give 0.8993 and random neighbors
0.8995, a difference of 0.02 points, whereas genuine retrieval reaches
0.9092. Because genuine and random retrieval use the same projectors
and token positions, added retrieval capacity alone does not explain
the true-versus-random gap. The 0.02-point random-versus-zeroed
difference is negligible relative to full-model seed variability and
is not interpreted as an effect of the token positions themselves.

A paired comparison across all 1{,}443 tasks gives a more
granular picture. PRISM exceeds the random-neighbor control on 701
tasks and falls below it on 633, with 109 ties, so it wins 52.5\% of
the 1{,}334 non-tied comparisons (two-sided exact binomial
$p = 0.067$). The macro-average moves by 0.97 points while the median
task moves very little, which locates the benefit of retrieved content
in a subset of tasks rather than spreading it evenly across the
nosology. The aggregate improvement is therefore concentrated in a subset of tasks.

Table~\ref{tab:ablation} presents the remaining component removal experiments.
\paragraph{Structured EHR encoding drives diagnostic performance.}
Removing the FT-Transformer costs 2.20 AUROC points, with the
largest loss on the diagnosis nosology and smaller changes on
deterioration and ICU admission. This pattern is consistent with many
diagnosis tasks drawing strongly on laboratory measurements such as
troponin, creatinine, and glucose. We therefore interpret the
structured record as an important complementary source of
discriminative information in PRISM.
\paragraph{Numerical soft-token pathway versus full text serialization.}
Replacing the FT-Transformer soft-token pathway with full
text-serialized EHR features reduces AUROC by 3.66 points, from 0.9092
to 0.8726, the largest tested substitution. The standard textual
patient summary is retained in both conditions; this ablation changes
the additional structured-EHR pathway rather than comparing numerical
input against the absence of text.
A 2{,}048-token budget cannot contain all 470 variables once fully
serialized without truncation, so the text condition instead receives
the required 8{,}192-token maximum sequence length. Nothing is
truncated; at batch size four it processes 32{,}768 tokens per step,
versus 16{,}384 for PRISM at batch size eight. Thus truncation and
context starvation do not explain the observed gap. The comparison
does, however, necessarily differ in sequence length, batch size, and
optimization conditions, so we interpret it as a practical comparison
between full serialization and the compact numerical soft-token
pathway rather than a pure causal isolation of numerical
representation. The structured record occupies roughly four thousand
serialized tokens versus eight projected soft tokens, and inference on
identical hardware increases from approximately four hours to
twenty-four under full serialization.
\paragraph{ECG modalities contribute dissociable and complementary signal.}
Removing the current-visit ECG encoder costs 1.61 points and
removing longitudinal history costs 1.22, indicating complementary
waveform signal within PRISM.
Within the longitudinal stream, we compare recency weighting with
simpler pooling on the 122 encounters carrying two or more prior
acquisitions, the only population on which the rules differ. The
learned rule has the highest point estimate at 0.9136, compared with
0.9069 for the best tested fixed decay, 0.9026 for uniform averaging,
0.8998 for the most recent acquisition only, and 0.8903 for the oldest
only. The corresponding differences are 0.67, 1.10, 1.38, and 2.33
AUROC points. The converged $|\lambda|=0.9891$ remains within 1.1\%
of its initialization, so we interpret the result as evidence for the
tested recency-weighted pooling rule rather than for learnability of
the decay constant. Because only 1.9\% of encounters are affected and
these are single-run subgroup comparisons, we treat the differences
as descriptive; across the full test fold the six rules span only
0.09 points, below the 0.32-point full-model seed standard deviation.
Appendix~\ref{app:decay} reports both strata in full.
\paragraph{Population retrieval benefits from dual-modality indexing.}
Activating either the EHR or ECG memory bank alone changes little
relative to no retrieval in the single-run evaluation, whereas
supplying both genuine modalities produces the full retrieval gain.
In that evaluation, no retrieval gives 0.8904, EHR-only 0.8910,
ECG-only 0.8906, and full PRISM 0.9094. Using the three-seed reference, zeroed retrieval gives 0.8993, random neighbors 0.8995, and
genuine retrieval 0.9092. Each comparison uses its corresponding full-model reference. Because genuine and random retrieval use the same
retrieval architecture, the true-versus-random difference supports a
content-dependent effect. The observed benefit appears only when both
genuine modalities are supplied, but these controls do not establish
why the interaction occurs. No joint-similarity index or explicit
intersection over neighbor sets is computed, and comparison with an
explicit joint retrieval geometry remains future work.
Same-subject retrieval occurs in 0.0237\% of training queries at
$k{=}3$ and is impossible at evaluation because the folds are
patient-disjoint, making it unlikely to explain the observed test
benefit. Neighborhood size is a descriptive sensitivity result:
$k{=}3$ has the highest point estimate among $\{1,3,5\}$, without
establishing why the alternatives perform differently.
\begin{table}[t]
\centering
\captionsetup{font=scriptsize}
\caption{%
  \textbf{ECG encoder backbone ablation.}
  All other architectural components and hyperparameters are held
  fixed; only the waveform encoder varies. Best AUROC (\%) across
  both inference pathways is reported per task group.
  Values are per-task-group percentages from the original
  single-run evaluation, on the same scale as
  Table~\ref{tab:benchmark} and not on the macro-of-macro scale used
  in Tables~\ref{tab:nollm}, \ref{tab:streams}
  and~\ref{tab:ablation}.
  $\Delta$ denotes the gap between D-BETA and the next-best encoder.
  \textbf{Bold}: best per column.
}
\small
\renewcommand{\arraystretch}{1.25}
\begin{tabular}{@{}l cccc@{}}
\toprule
\textbf{ECG Encoder}
  & \textbf{Diagnosis}
  & \textbf{Deterioration}
  & \textbf{ICU}
  & \textbf{Mortality} \\
\midrule
HeartBERT~\citep{tahery2024heartbert}
  & 83.65 & 91.07 & 90.79 & 91.61 \\
ECGFM-KED~\citep{tian2024ked}
  & 82.71 & 90.57 & 90.27 & 90.90 \\
ECGFounder~\citep{li2024ecgfounder}
  & 82.54 & 91.71 & 90.60 & 90.73 \\
\midrule
\rowcolor{highlightgreen}
\textbf{D-BETA (PRISM)}~\citep{hung2025boosting}
  & \textbf{86.24} & \textbf{93.12} & \textbf{91.79} & \textbf{92.82} \\
\rowcolor{highlightgreen!15}
\quad $\Delta$ \textit{vs.\ next-best}
  & \small{+2.59} & \small{+1.41} & \small{+1.00} & \small{+1.21} \\
\bottomrule
\end{tabular}
\label{tab:ecg_encoder_ablation}
\end{table}
\paragraph{Training-stage ablations.}
Freezing the retrieval projectors costs 3.90 AUROC points and disabling LoRA adaptation costs 8.75, the largest single change among the tested ablations. These results show that trainable retrieval alignment and LoRA adaptation are both important in the tested PRISM configuration; they do not by themselves identify a unique failure mechanism.
\paragraph{ECG encoder selection.}
Table~\ref{tab:ecg_encoder_ablation} evaluates three alternative ECG encoder backbones within PRISM with the remaining components held fixed. D-BETA has the highest point estimate on every task group, with margins of +2.59 points on Diagnosis and +1.41 on Deterioration over the next-best encoder. D-BETA's joint waveform--text pre-training is compatible with the retrieval setting, but this ablation varies the encoder backbone as a whole and does not isolate why D-BETA performs better.
\subsection{Control Summary}
\label{sec:summary}
Random-neighbor and zeroed retrieval differ by only 0.02
AUROC points, below the 0.32-point full-model seed standard deviation,
while genuine retrieval is higher than both. Neighborhood size and
history pooling also show sensitivity to the tested choices, with
$k{=}3$ and the learned decay rule giving the highest point estimates
among their respective alternatives. Because $k$ was selected on
held-out performance, $\lambda$ was learned, and the pooling
comparisons affect only a small subgroup, we treat these patterns as
descriptive controls rather than causal evidence.

\subsection{Additional Metrics and Subgroup Analysis}
\label{sec:additional_metrics}
Many of the 1{,}443 tasks are substantially imbalanced, so AUROC
alone does not fully characterize performance, and we therefore report
AUPRC and calibration alongside it.
Table~\ref{tab:permetric} reports discrimination, average
precision and calibration for each task group under both inference
pathways. Average precision appears twice, at the retained evaluation prevalence after ECG exclusions and corrected to true cohort prevalence, since only the
latter is clinically interpretable. The two pathways are close
throughout, with the generative pathway slightly ahead on aggregate,
and calibration is best on the ICU and mortality endpoints and weakest
on the diagnosis nosology.

\begin{table}[t]
\centering
\captionsetup{font=scriptsize}
\caption{\textbf{Per-group discrimination, average precision and
calibration.} AP@eval is measured at the retained evaluation prevalence after ECG exclusions; its chance reference is the retained positive fraction for each task.
AP@true rescales to the cohort prevalence of each endpoint. ECE uses
ten equal-count bins weighted by bin size.}
\label{tab:permetric}
\small
\renewcommand{\arraystretch}{1.25}
\begin{tabular}{@{}lrrccccc@{}}
\toprule
\textbf{Group} & \textbf{Tasks} & \textbf{Rows} & \textbf{AUROC} & \textbf{AP@eval} & \textbf{AP@true} & \textbf{Brier} & \textbf{ECE} \\
\midrule
\multicolumn{8}{@{}l}{\textit{PRISM}$_{\textit{gen}}$} \\
Diagnoses     & 1{,}428 & 253{,}055 & 0.8619 & 0.8532 & 0.3400 & 0.1720 & 0.1558 \\
Deterioration & 6       & 729       & 0.9291 & 0.9357 & 0.5855 & 0.1186 & 0.1183 \\
ICU           & 2       & 3{,}476   & 0.9180 & 0.9204 & 0.7265 & 0.1383 & 0.0954 \\
Mortality     & 7       & 4{,}898   & 0.9279 & 0.9172 & 0.5495 & 0.1126 & 0.0840 \\
\rowcolor{highlightgreen}
\textbf{Macro-of-macro} & 1{,}443 & & \textbf{0.9092} & & \textbf{0.5504} & & \textbf{0.1134} \\
\midrule
\multicolumn{8}{@{}l}{\textit{PRISM}$_{\textit{disc}}$} \\
Diagnoses     & 1{,}428 & 253{,}055 & 0.8599 & 0.8499 & 0.3232 & 0.2015 & 0.1875 \\
Deterioration & 6       & 729       & 0.9293 & 0.9371 & 0.5750 & 0.1356 & 0.1319 \\
ICU           & 2       & 3{,}476   & 0.9160 & 0.9077 & 0.6642 & 0.1628 & 0.1367 \\
Mortality     & 7       & 4{,}898   & 0.9272 & 0.9153 & 0.5551 & 0.1331 & 0.1132 \\
\rowcolor{highlightgreen!40}
\textbf{Macro-of-macro} & 1{,}443 & & \textbf{0.9081} & & \textbf{0.5294} & & \textbf{0.1423} \\
\bottomrule
\end{tabular}
\end{table}
We report subgroup discrimination and calibration in
Appendix~\ref{app:subgroup}, but we do not evaluate calibration under
temporal or cross-site distribution shift. AUROC declines across the
age bands from 0.9277 in patients under 40 to 0.8684 in those aged 85
and over. Twenty of the twenty-one pairwise subgroup interval
comparisons overlap; the exception is the 40--54 versus 85+ age bands,
and no sex or self-reported ethnicity comparison separates.
Excluding the Hispanic/Latino subgroup, subgroup AUROC is associated with subgroup event rate ($r=-0.78$), as detailed in Appendix~\ref{app:subgroup}. This descriptive association is consistent with differences in case mix but does not establish their cause. We therefore treat these analyses
as descriptive rather than as evidence of demographic equivalence or
disparity.

\section{Discussion and Conclusion}
\label{sec:discussion}

Across the evaluations, PRISM's performance reflects both the evidence available to the model and how that evidence is integrated. Matching the benchmark observation window brings PRISM close to prior multimodal systems, while comparisons with the tested compact fusion models favor PRISM under matched inputs. On MC-MED, supervised fine-tuning contributes a larger improvement than replacing the memory bank alone. Together, these findings motivate evaluating representation, observation window, and local adaptation as distinct design choices.

The retrieval controls further distinguish the presence of context from the selection of useful neighbors. Genuine retrieval exceeds random retrieval by 0.97 AUROC points on MDS-ED, 0.05 points in the temporal evaluation, and 3.31 points after MC-MED fine-tuning. These evaluations differ in cohort, eligible tasks, and training configuration, so the contrasts do not isolate the cause of this variation. Their practical implication is to assess retrieved content against random and zeroed controls in the intended setting, rather than assuming that a memory bank will improve performance.

PRISM extends retrieval augmentation from text corpora to learned
representations of training-cohort encounters. In the deployed model,
only mean-pooled embeddings enter the language model; retrieved-neighbor
labels and individual records do not. The underlying neighbors remain
enumerable offline, providing retrieval provenance for retrospective
analysis without exposing individual records through the prompt. How
such provenance should be presented to end users remains an open
deployment question.
\paragraph{Limitations.}
The evaluation remains retrospective, spans two institutions, and uses a single language-model backbone, MedGemma-4B. Broader multi-site and prospective evaluation is needed to characterize robustness to acquisition, measurement, and population differences. The strongest external result uses supervised local adaptation, and the temporal baseline comparisons are restricted to the tested compact fusion architectures. Subgroup analyses in Appendix~\ref{app:subgroup} describe performance within the evaluated cohort, while calibration under distribution shift and the clinical use of retrieved evidence remain to be assessed.

\paragraph{Conclusion.}
PRISM integrates numerical clinical features, ECG waveforms, longitudinal history, and retrieved encounter representations within a multimodal language model. The MDS-ED experiments and supervised MC-MED adaptation support our approach to heterogeneous evidence integration, while the matched-window and retrieval controls identify where additional inputs and selected neighbor representations contribute. These findings motivate evaluation of the complete evidence pipeline when adapting multimodal models to new clinical settings.
\subsection*{Broader Impact Statement}
Retrieving over a patient population rather than a curated corpus
introduces two classes of risk, and both concern deployment rather
than the retrospective evaluation reported here. The first is
disclosure. PRISM passes only mean-pooled retrieval embeddings into the language
model, and no retrieved-neighbor attribute, identifier, discharge code
or outcome is surfaced through the retrieval pathway, which limits
direct exposure of individual neighbor records. That property belongs to the current interface rather than to
the method, and it would not survive a deployment that displayed
individual retrieved cases in order to justify a prediction. Such an
interface would expose patient attributes and documented outcomes
directly, and it would raise the risk of re-identification through
rare combinations of clinical findings, since a neighbor matched
closely on an unusual presentation may be effectively unique in the
source cohort. Membership inference remains a consideration even where
only latent vectors are retained, because an index built over real
encounters may reveal whether a given patient contributed to it. A
deployment that surfaces retrieved cases would therefore require
privacy analysis that this work does not provide.
The second risk is interpretive. Similarity in a learned embedding
space is not evidence of shared causation. Neighbors who resemble a
patient on measured features may differ in treatment history,
unrecorded comorbidity, demographic composition or site-specific
practice, and presenting them as support for a prediction may
encourage clinicians to reason by analogy from cases that are alike
only in the respects the encoder happens to represent. We would regard
clinical deployment as premature until subgroup performance has been
characterized, until calibration has been assessed under distribution
shift rather than on a matched retrospective split, and until the
framing of retrieved evidence to end users has been evaluated
prospectively.

\subsection*{Acknowledgments}
This work was supported by the NWO AiNed Fellowship Grant of A.S., and in part by Google.org and the Google Cloud Research Credits program through the Gemini Academic Program. We also acknowledge the use of the Dutch National Supercomputer Snellius for essential computational tasks.

\bibliography{main}
\bibliographystyle{tmlr}
\newpage
\appendix
\section*{Appendix}
\begin{enumerate}[label=\Alph*.,leftmargin=2em,itemsep=2pt]
  \item Training Hyperparameters \dotfill \pageref{app:hyperparams}
  \item Soft-Token Projector Specifications \dotfill \pageref{app:proj}
  \item Population Memory Bank Configuration \dotfill \pageref{app:memory_bank}
  \item Per-Task AUROC: Deterioration, ICU, and Mortality \dotfill \pageref{app:pertask}
  \item Diagnosis AUROC by ICD-10 Chapter \dotfill \pageref{app:icd}
  \item AUROC Distribution Across Diagnosis Tasks \dotfill \pageref{app:auroc_dist}
  \item Longitudinal ECG History: Availability and Aggregation \dotfill \pageref{app:decay}
  \item Cohort and Dataset Details \dotfill \pageref{app:cohort}
  \item Performance by Patient Subgroup \dotfill \pageref{app:subgroup}
  \item Clinical Prompt Design and Representative Examples \dotfill \pageref{app:prompt}
  \item Training Data Construction and Prompt Generation \dotfill \pageref{app:data_construction}
  \item Population Memory Bank: Retrieval Across Clinical Phenotypes \dotfill \pageref{app:retrieval_phenotypes}
  \item External and Temporal Evaluation Protocols \dotfill \pageref{app:external_temporal_protocol}
  \item Temporal Generalization \dotfill \pageref{app:temporal_generalization}
\end{enumerate}
\section{Training Hyperparameters}
\label{app:hyperparams}
PRISM is trained in two stages with distinct sets of trainable parameters. Stage~1 jointly optimizes the EHR FT-Transformer, the temporal history aggregator (including the learnable decay scalar $\lambda$), and the three primary soft-token projectors (ECG, EHR, history), while the MedGemma-4B backbone and D-BETA encoder remain frozen. Stage~2 freezes all Stage~1 parameters and jointly trains the LoRA adapters, the two retrieval projectors (initialized from scratch), and the classification head (loaded from the Stage~1 checkpoint). Table~\ref{tab:hyper} reports all hyperparameters for both stages.
\begin{table}[h]
\centering
\small
\caption{Training hyperparameters for PRISM Stage~1 and Stage~2.}
\label{tab:hyper}
\setlength{\tabcolsep}{6pt}
\renewcommand{\arraystretch}{1.25}
\begin{tabular}{lcc}
\toprule
\textbf{Hyperparameter} & \textbf{Stage 1} & \textbf{Stage 2} \\
\midrule
LLM backbone & MedGemma-4B (frozen) & MedGemma-4B + LoRA \\
LoRA rank / scaling & -- & 16 / 32 \\
LoRA dropout & -- & 0.05 \\
LoRA target modules & -- & q, k, v, o, gate, up, down \\
\midrule
EHR encoder $d_{\mathrm{model}}$ & 192 & frozen \\
EHR encoder layers / heads & 3 / 4 & frozen \\
EHR embedding $d_{\mathrm{repr}}$ & 512 & frozen \\
History decay $\lambda$ (init) & 1.0 & frozen \\
\midrule
Soft tokens per modality $n_m$ & 8 & 8 \\
Projector hidden dim & 1{,}024 & 1{,}024 \\
Classification weight $\alpha$ & 0.1 & 0.1 \\
Memory bank $k$ & -- & 3 \\
\midrule
Effective batch size & 32 (micro 4 $\times$ accum 8) & 2 \\
Learning rate & $2 \times 10^{-4}$ & $2 \times 10^{-5}$ \\
LR schedule & Cosine (100 warm-up) & Cosine \\
Gradient clipping & $\|\mathbf{g}\|_2 \leq 1.0$ & $\|\mathbf{g}\|_2 \leq 1.0$ \\
Max sequence length & 2{,}048 tokens & 2{,}048 tokens \\
Mixed precision & BF16 & BF16 \\
Hardware & 2$\times$ H100 80\,GB & 2$\times$ H100 80\,GB \\
\bottomrule
\end{tabular}
\end{table}
\section{Soft-Token Projector Specifications}
\label{app:proj}
Each modality embedding is mapped to a sequence of eight soft tokens in the language model's embedding space through a two-layer feed-forward projector with layer normalization, GELU activation, and a final linear expansion (Equation~\ref{eq:softtok}). All five projectors share an identical architecture and differ only in input dimensionality and training stage. The three primary projectors (current ECG, EHR, ECG history) are trained during Stage~1. The two retrieval projectors are initialized from scratch and trained during Stage~2 alongside the LoRA adapters. Table~\ref{tab:proj} summarizes the specifications.
\begin{table}[h]
\centering
\small
\caption{Soft-token projector specifications. All projectors have a hidden dimension of 1{,}024 and produce an output shape of $8 \times 2{,}560$ (MedGemma-4B embedding size).}
\label{tab:proj}
\setlength{\tabcolsep}{5pt}
\renewcommand{\arraystretch}{1.25}
\begin{tabular}{lcc}
\toprule
\textbf{Modality stream} & \textbf{Input dim.} & \textbf{Stage} \\
\midrule
Current-visit ECG & $d_{\mathrm{ecg}}$ & 1 \\
Structured EHR & 512 & 1 \\
Longitudinal ECG history & 512 & 1 \\
Retrieved EHR context & 512 & 2 \\
Retrieved ECG context & $d_{\mathrm{ecg}}$ & 2 \\
\bottomrule
\end{tabular}
\end{table}
\section{Population Memory Bank Configuration}
\label{app:memory_bank}
The population memory bank comprises two independent nearest-neighbor indices constructed over per-encounter embeddings from the 116{,}186 training-fold encounters after Stage~1 training. The EHR index operates over 512-dimensional EHR embeddings and the ECG index over $d_{\mathrm{ecg}}$-dimensional ECG embeddings. Both use $\ell_2$-normalized vectors with cosine similarity as the distance metric.
To eliminate retrieval latency during training and inference, neighbor assignments for all encounters are materialized as pre-computed row-aligned lookup tables immediately after index construction. At runtime, each retrieval operation reduces to a single array index per modality. The top-$k$ retrieved embeddings from each index are independently mean-pooled to produce context vectors before projection into the language model's token space. Table~\ref{tab:memory_bank} summarizes the configuration.
\begin{table}[h]
\centering
\small
\caption{Population memory bank configuration for PRISM.}
\label{tab:memory_bank}
\setlength{\tabcolsep}{8pt}
\renewcommand{\arraystretch}{1.25}
\begin{tabular}{lcc}
\toprule
\textbf{Property} & \textbf{EHR Index} & \textbf{ECG Index} \\
\midrule
Vector dimension & 512 & $d_{\mathrm{ecg}}$ \\
Distance metric & Cosine similarity & Cosine similarity \\
Indexed encounters & 116{,}186 & 116{,}186 \\
Neighbors retrieved ($k$) & 3 & 3 \\
Context aggregation & Mean pooling & Mean pooling \\
Index construction & After Stage~1 & After Stage~1 \\
\bottomrule
\end{tabular}
\end{table}
\section{Per-Task AUROC: Deterioration, ICU, and Mortality}
\label{app:pertask}
Table~\ref{tab:pertask} reports per-task AUROC for all 15 non-diagnosis tasks on the held-out test set, together with 95\% confidence intervals computed via empirical bootstrap (1{,}000 resamples per task). PRISM achieves AUROC above 0.95 on five tasks, including cardiac arrest, mechanical ventilation, and within-1-day mortality, and maintains AUROC above 0.90 on 11 of 15 tasks. Performance decreases as the mortality time horizon extends, consistent with the increasing contribution of chronic disease burden over longer intervals relative to the acute physiological state captured in the emergency department window.
\begin{table}[h]
\centering
\small
\caption{Per-task AUROC for all Deterioration, ICU, and Mortality tasks on the held-out test set, with 95\% bootstrap confidence intervals (1{,}000 resamples). The stronger pathway per task is \textbf{bolded}. Point estimates are computed on the retained sampled evaluation records; the bootstrap estimates uncertainty. The macro-averages in the final three rows are the exact means of the per-task values above them and reconcile with the corresponding entries of Table~\ref{tab:benchmark}.}
\label{tab:pertask}
\setlength{\tabcolsep}{4pt}
\renewcommand{\arraystretch}{1.25}
\begin{tabular}{llcll}
\toprule
\textbf{Group} & \textbf{Endpoint} & \textbf{Test $n$} & $\textbf{PRISM}_{\textbf{disc}}$ \textbf{[95\% CI]} & $\textbf{PRISM}_{\textbf{gen}}$ \textbf{[95\% CI]} \\
\midrule
\multirow{6}{*}{Deterioration}
  & Cardiac arrest & 68 & 0.9706~[0.9187, 0.9965] & \textbf{0.9870}~[0.9627, 1.0000] \\
  & Inotrope administration & 27 & \textbf{0.9615}~[0.8765, 1.0000] & 0.9286~[0.8076, 1.0000] \\
  & Mechanical ventilation & 437 & 0.9597~[0.9438, 0.9759] & \textbf{0.9666}~[0.9517, 0.9789] \\
  & ECMO cannulation & 18 & \textbf{0.9383}~[0.8051, 1.0000] & \textbf{0.9383}~[0.7916, 1.0000] \\
  & Vasopressor initiation & 125 & \textbf{0.9286}~[0.8717, 0.9701] & 0.9247~[0.8742, 0.9672] \\
  & Severe hypoxemia & 54 & \textbf{0.8285}~[0.7069, 0.9293] & \textbf{0.8285}~[0.7129, 0.9286] \\
\midrule
\multirow{2}{*}{ICU}
  & Admission within 24h & 1{,}588 & 0.9191~[0.9061, 0.9329] & \textbf{0.9221}~[0.9100, 0.9349] \\
  & Admission during stay & 1{,}888 & 0.9118~[0.8987, 0.9239] & \textbf{0.9136}~[0.9002, 0.9249] \\
\midrule
\multirow{7}{*}{Mortality}
  & During hospitalization & 18 & \textbf{1.0000}~[1.0000, 1.0000] & \textbf{1.0000}~[1.0000, 1.0000] \\
  & Within 1 day & 97 & \textbf{0.9864}~[0.9574, 0.9983] & 0.9817~[0.9540, 0.9979] \\
  & Within 7 days & 245 & 0.9404~[0.9101, 0.9669] & \textbf{0.9421}~[0.9086, 0.9680] \\
  & Within 28 days & 535 & 0.9188~[0.8939, 0.9414] & \textbf{0.9224}~[0.8993, 0.9442] \\
  & Within 90 days & 922 & 0.8970~[0.8762, 0.9150] & \textbf{0.8971}~[0.8772, 0.9158] \\
  & Within 180 days & 1{,}268 & 0.8759~[0.8573, 0.8950] & \textbf{0.8761}~[0.8561, 0.8928] \\
  & Within 365 days & 1{,}813 & 0.8771~[0.8625, 0.8915] & \textbf{0.8780}~[0.8629, 0.8926] \\
\midrule
\multicolumn{2}{l}{Deterioration macro-avg (6 tasks)} & & 0.9312~[0.8958, 0.9541] & 0.9290~[0.8916, 0.9582] \\
\multicolumn{2}{l}{ICU macro-avg (2 tasks)} & & 0.9155~[0.9070, 0.9247] & 0.9179~[0.9092, 0.9265] \\
\multicolumn{2}{l}{Mortality macro-avg (7 tasks)} & & 0.9279~[0.9191, 0.9347] & 0.9282~[0.9198, 0.9350] \\
\bottomrule
\end{tabular}
\end{table}
\section{Diagnosis AUROC by ICD-10 Chapter}
\label{app:icd}
Table~\ref{tab:icd_chapter} aggregates AUROC across all 1{,}428 diagnosis tasks by ICD-10 clinical chapter under the generative pathway, sorted by mean AUROC. PRISM attains mean AUROC above 0.87 in 11 of 23 chapters and above 0.80 in all chapters. The highest-performing chapters are those in which specific laboratory abnormalities are directly pathognomonic: obstetric conditions (Chapter~O, mean 0.98), infectious diseases (Chapter~A, 0.92), and circulatory disorders (Chapter~I, 0.89). Performance is comparatively lower in chapters dominated by symptom-based or musculoskeletal diagnoses, where the primary discriminating information resides in clinical presentation rather than the structured biochemical record.
\begin{table}[h]
\centering
\small
\caption{Diagnosis AUROC by ICD-10 chapter ($\text{PRISM}_{\text{gen}}$), sorted by mean AUROC. $N$: number of ICD-10 codes per chapter.}
\label{tab:icd_chapter}
\setlength{\tabcolsep}{5pt}
\renewcommand{\arraystretch}{1.25}
\begin{tabular}{llccc}
\toprule
\textbf{ICD-10 Chapter} & \textbf{Ch.} & $\boldsymbol{N}$ & \textbf{Mean} & \textbf{Range} \\
\midrule
Pregnancy, childbirth, puerperium & O & 3 & 0.9825 & 0.9475--1.0000 \\
Certain infectious diseases & A & 15 & 0.9189 & 0.8493--0.9835 \\
Blood and immune disorders & D & 50 & 0.8952 & 0.6600--1.0000 \\
Circulatory system & I & 181 & 0.8900 & 0.7128--1.0000 \\
Poisoning and toxic effects & T & 43 & 0.8888 & 0.6317--0.9889 \\
Neoplasms & C & 48 & 0.8886 & 0.2800--1.0000 \\
External causes (V) & V & 38 & 0.8838 & 0.7376--0.9698 \\
Factors influencing health & Z & 166 & 0.8788 & 0.5091--1.0000 \\
Infectious and parasitic (B) & B & 37 & 0.8766 & 0.7381--0.9835 \\
External causes (W) & W & 11 & 0.8732 & 0.7784--0.9705 \\
Endocrine, nutritional, metabolic & E & 107 & 0.8703 & 0.6875--1.0000 \\
Skin and subcutaneous & L & 25 & 0.8681 & 0.6562--0.9835 \\
Digestive system & K & 119 & 0.8670 & 0.6942--1.0000 \\
Genitourinary system & N & 39 & 0.8660 & 0.6730--0.9952 \\
External causes (Y) & Y & 27 & 0.8577 & 0.7427--0.9917 \\
Respiratory system & J & 71 & 0.8519 & 0.5091--0.9905 \\
Injury and trauma & S & 43 & 0.8517 & 0.6317--0.9877 \\
Nervous system & G & 67 & 0.8431 & 0.6159--0.9738 \\
Mental and behavioral & F & 71 & 0.8411 & 0.5102--0.9982 \\
Congenital malformations & Q & 3 & 0.8411 & 0.7842--0.9224 \\
Symptoms, signs, abnormal findings & R & 171 & 0.8235 & 0.5091--0.9961 \\
Eye and ear & H & 24 & 0.8232 & 0.6311--0.9506 \\
Musculoskeletal system & M & 68 & 0.8038 & 0.4201--1.0000 \\
\midrule
\textbf{Overall (1{,}428 tasks)} & & \textbf{1{,}428} & \textbf{0.8624} & 0.2800--1.0000 \\
\bottomrule
\end{tabular}
\end{table}
\section{AUROC Distribution Across Diagnosis Tasks}
\label{app:auroc_dist}
Table~\ref{tab:auroc_dist} stratifies all 1{,}428 diagnosis tasks by AUROC range under the discriminative pathway. The majority of tasks (78.4\%) exceed AUROC 0.80, and more than one third (35.4\%) exceed 0.90. Only eight tasks (0.6\%) fall below 0.60, all involving symptom-based or psychiatric diagnoses where the discriminating signal is largely absent from the structured laboratory and waveform record.
\begin{table}[h]
\centering
\small
\caption{Distribution of per-task AUROC across all 1{,}428 diagnosis tasks ($\text{PRISM}_{\text{disc}}$). Cumulative percentage is computed from the highest bin downward.}
\label{tab:auroc_dist}
\setlength{\tabcolsep}{6pt}
\renewcommand{\arraystretch}{1.25}
\begin{tabular}{lccc}
\toprule
\textbf{AUROC Range} & \textbf{Tasks} & \textbf{Proportion} & \textbf{Cumulative} \\
\midrule
$\geq 0.90$ & 506 & 35.4\% & 35.4\% \\
$0.80$--$0.89$ & 613 & 42.9\% & 78.4\% \\
$0.70$--$0.79$ & 260 & 18.2\% & 96.6\% \\
$0.60$--$0.69$ & 41 & 2.9\% & 99.4\% \\
$< 0.60$ & 8 & 0.6\% & 100\%  \\
\midrule
\textbf{Total} & \textbf{1{,}428} & \textbf{100\%} & \\
\bottomrule
\end{tabular}
\end{table}
\section{Longitudinal ECG History: Availability and Aggregation}
\label{app:decay}
Table~\ref{tab:histavail} reports how many prior ECG acquisitions
each test encounter has within the 96-hour window. The distribution is
strongly skewed, and it determines what the temporal aggregator can
influence. For encounters with no prior acquisition the learned
no-history embedding is used in place of $\mathbf{z}_{\mathrm{hist}}$.
For encounters with exactly one, the normalized weight of
Equation~\ref{eq:decay} is $1$ regardless of $\lambda$. The
aggregation rule therefore alters the input for only the 122
encounters with two or more acquisitions, or 1.9\% of the test fold,
which bounds how far any two aggregation rules can differ on a
macro-averaged metric.

\begin{table}[h]
\centering
\small
\caption{Prior ECG acquisitions per test encounter within the 96-hour window.}
\label{tab:histavail}
\setlength{\tabcolsep}{10pt}
\renewcommand{\arraystretch}{1.25}
\begin{tabular}{lrr}
\toprule
\textbf{Prior ECGs} & \textbf{Encounters} & \textbf{Share} \\
\midrule
0 & 5{,}752 & 89.33\% \\
1 & 565     & 8.77\% \\
2 & 86      & 1.34\% \\
3 & 26      & 0.40\% \\
4 & 7       & 0.11\% \\
5 & 2       & 0.03\% \\
6 & 1       & 0.02\% \\
\midrule
$\geq 2$ (aggregator operative) & 122 & 1.89\% \\
\textbf{Total} & \textbf{6{,}439} & \textbf{100\%} \\
\bottomrule
\end{tabular}
\end{table}

Table~\ref{tab:decayparam} reports the converged decay parameter
and the quantities derived from it. Because $|\lambda|$ remains within
1.1\% of its initialization, the ratio $\tau/|\lambda|$ returns
approximately $\tau$ by construction and should not be read as a
timescale recovered from data.

\begin{table}[h]
\centering
\small
\caption{Converged temporal decay parameter and derived quantities.}
\label{tab:decayparam}
\setlength{\tabcolsep}{10pt}
\renewcommand{\arraystretch}{1.25}
\begin{tabular}{lr}
\toprule
\textbf{Quantity} & \textbf{Value} \\
\midrule
Initialization of $\lambda$ & 1.0000 \\
Converged $|\lambda|$ & 0.9891 \\
Change from initialization & 1.1\% \\
Half-life $\tau \ln 2 / |\lambda|$ & 67.3 h \\
$\tau / |\lambda|$ & 97.1 h \\
Weight retained at 96 h & 0.372 \\
Weights at 0.4 h and 0.6 h & 0.5005 / 0.4995 \\
\bottomrule
\end{tabular}
\end{table}

Table~\ref{tab:aggvariants} compares the learned rule against
fixed and uniform alternatives under both strata. The right-hand
column restricts evaluation to the 122 encounters carrying two or more
prior acquisitions, which is the only population on which the rules
produce different inputs. The left-hand column covers the full test
fold. Its narrow 0.09-point range is the arithmetic consequence of a
2.33-point effect confined to 1.9\% of encounters, and it falls below
the 0.32-point seed standard deviation of the full model.

\begin{table}[h]
\centering
\small
\caption{Alternative rules for pooling prior ECG embeddings.}
\label{tab:aggvariants}
\setlength{\tabcolsep}{8pt}
\renewcommand{\arraystretch}{1.25}
\begin{tabular}{lcc}
\toprule
\textbf{Aggregation rule} & \textbf{All encounters} & \textbf{$\geq 2$ prior ECGs} \\
\midrule
Oldest acquisition only      & 0.9083 & 0.8903 \\
Most recent acquisition only & 0.9087 & 0.8998 \\
Uniform average              & 0.9091 & 0.9026 \\
Fixed decay, half-life 1 h   & 0.9090 & 0.9041 \\
Fixed decay, half-life 2 h   & 0.9089 & 0.9069 \\
\rowcolor{highlightgreen}
\textbf{Learned decay (PRISM)} & \textbf{0.9092} & \textbf{0.9136} \\
\bottomrule
\end{tabular}
\end{table}

\section{Cohort and Dataset Details}
\label{app:cohort}
PRISM is trained and evaluated on the MDS-ED benchmark~\citep{lopezalcaraz2025mdsed}, derived from the MIMIC-IV and MIMIC-IV-ECG databases~\citep{johnson2023mimic}. This dataset constitutes the largest publicly available source of linked ECG waveforms and structured EHR data for emergency department encounters. We adopt the standardized 20-fold stratified splits defined by the MDS-ED protocol to ensure direct comparability with all existing baselines.
To represent the information available to a clinician at the point of disposition, all benchmark data is extracted from a 90-minute window beginning at ED arrival. Every encounter must have a raw 12-lead ECG, at least one structured laboratory measurement, and a complete discharge record with ICD-10-CM codes. The feature vector ($D{=}470$) includes demographics, biometrics, triage vitals, and laboratory panels; previous medications and free-text complaints are excluded.
PRISM additionally consumes a retrospective 96-hour window ending
at ED arrival, and the supplementary vital sign statistics of
Section~\ref{sec:ehr_enc} and the prior ECG acquisitions of
Section~\ref{sec:history} are drawn from it. This window lies entirely
before the benchmark window and therefore entirely before the
prediction cutoff, so it introduces no post-cutoff information. It
does give PRISM access to history that the MDS-ED and UniPACT
baselines did not receive; the full-input benchmark comparison therefore
includes this input-window difference.
\paragraph{Evaluation protocol.}
The test-set builder first retains all positives and an equal random sample of negatives for each task, yielding 265{,}816 exactly balanced encounter--task records. Records from 83 encounters with unavailable ECG waveforms are subsequently excluded: 81 encounters are absent from the MIMIC-IV-ECG index and two have unreadable waveforms. The retained evaluation contains 262{,}158 records and is not necessarily exactly balanced within each task. Initial class-balanced sampling follows the benchmark convention used by UniPACT~\citep{tang2026unipact}; ECG-based exclusions are an additional selection step. AUROC measures positive--negative ranking, whereas average precision and calibration depend on the evaluation prevalence. We distinguish metrics on the retained sampled records from metrics adjusted to cohort prevalence.
MDS-ED is a multi-label
architecture rather than a per-task binary predictor, which is a
structural difference between the systems and not a difference in
evaluation.
The benchmark test fold contains 6{,}454 task-eligible encounters, of
which 6{,}439 have a resolvable prior-ECG history and are used in
Appendix~\ref{app:decay}, and 6{,}356 carry complete demographic
coding and are used in Appendix~\ref{app:subgroup}. Following the
benchmark protocol, one ECG per stay is used.
The Brier score and expected calibration error reported here describe calibration on the retained sampled evaluation records after ECG exclusions, rather than on an exactly balanced set or at natural cohort prevalence. They are
therefore comparable across the systems and ablations reported here,
which all use the same protocol, but should not be read as the
calibration a deployed model would exhibit on an unselected ED
population.
Aggregate figures are computed as a macro-of-macro: per-task AUROC is
averaged within each of the four task groups, and the four group means
are then averaged. This prevents the 1{,}428 diagnosis tasks from
dominating the fifteen deterioration, ICU and mortality endpoints,
which are the outcomes of greatest clinical consequence. All reported
systems use the same rule. Under task-weighted averaging the
corresponding figures are 0.8626 for AUROC and 0.3426 for average
precision at true prevalence.

The MDS-ED stratified folds are disjoint at the level of the
patient, so no patient contributes encounters to more than one
partition. The population memory bank is built only from training-fold
encounters, so a validation or test query cannot return an encounter
belonging to the same patient, and self-retrieval is precluded by
construction. The same guarantee does not extend within the training
fold, where a patient with multiple encounters may retrieve one of
their own. Section~\ref{sec:memory_bank} discusses the consequences.
Table~\ref{tab:cohort_stats} summarizes the cohort statistics and experimental partitions. Table~\ref{tab:clinical_groups} describes the four clinical task groups predicted by PRISM.
\begin{table}[h]
\centering
\small
\caption{MDS-ED cohort and partition statistics.}
\label{tab:cohort_stats}
\setlength{\tabcolsep}{10pt}
\renewcommand{\arraystretch}{1.25}
\begin{tabular}{lr}
\toprule
\textbf{Metric} & \textbf{Value} \\
\midrule
Unique patients & 71{,}098 \\
ECG-linked encounters & 129{,}095 \\
Training samples (task instances) & 4.5M \\
\midrule
Training set & 116{,}186 \\
Validation set & 6{,}455 \\
Test set & 6{,}454 \\
\bottomrule
\end{tabular}
\end{table}
\begin{table}[h]
\centering
\small
\caption{Summary of the 1{,}443 binary clinical tasks predicted by PRISM.}
\label{tab:clinical_groups}
\setlength{\tabcolsep}{6pt}
\renewcommand{\arraystretch}{1.25}
\begin{tabular}{lcl}
\toprule
\textbf{Group} & \textbf{Tasks} & \textbf{Examples / Endpoints} \\
\midrule
Diagnosis & 1{,}428 & Full ICD-10 nosology (circulatory, musculoskeletal, etc.) \\
Deterioration & 6 & Intubation, cardiac arrest, vasopressors, severe hypoxemia \\
ICU Admission & 2 & Overall ICU admission, admission within 24 hours \\
Mortality & 7 & In-hospital; within 1, 7, 28, 90, 180, 365 days \\
\bottomrule
\end{tabular}
\end{table}
\section{Performance by Patient Subgroup}
\label{app:subgroup}
Table~\ref{tab:subgroup} reports discrimination and calibration
for every demographic subgroup recorded in the cohort, with 95\%
confidence intervals obtained by bootstrap over patients. Groups are
ordered within each attribute by AUROC. Event rate is the proportion
of positive instances in the retained sampled evaluation set for that
subgroup, and differences in it reflect differences in case mix.

\begin{table}[h]
\centering
\small
\caption{PRISM performance stratified by sex, age band and
self-reported ethnicity. Confidence intervals are bootstrap estimates
over patients. Of the twenty-one pairwise contrasts within attributes,
twenty have overlapping intervals; the exception is the 40--54 band
against the 85+ band.}
\label{tab:subgroup}
\setlength{\tabcolsep}{4pt}
\renewcommand{\arraystretch}{1.25}
\begin{tabular}{llrrrcc}
\toprule
\textbf{Attribute} & \textbf{Subgroup} & \textbf{Patients} & \textbf{Rows} & \textbf{Event rate} & \textbf{AUROC [95\% CI]} & \textbf{ECE} \\
\midrule
Sex & Female & 3{,}329 & 132{,}751 & 0.478 & 0.9177 [0.9037, 0.9270] & 0.1143 \\
    & Male   & 3{,}027 & 129{,}407 & 0.523 & 0.8974 [0.8738, 0.9125] & 0.1442 \\
\midrule
Age & $<$40  & 1{,}028 & 30{,}289  & 0.265 & 0.9277 [0.8867, 0.9481] & 0.0962 \\
    & 40--54 & 1{,}073 & 38{,}031  & 0.416 & 0.9240 [0.9054, 0.9363] & 0.1153 \\
    & 55--69 & 1{,}938 & 82{,}045  & 0.522 & 0.8969 [0.8745, 0.9141] & 0.1378 \\
    & 70--84 & 1{,}598 & 76{,}755  & 0.565 & 0.8868 [0.8602, 0.9064] & 0.1356 \\
    & 85+    & 719     & 35{,}038  & 0.599 & 0.8684 [0.8253, 0.8918] & 0.1728 \\
\midrule
Ethnicity & Black / African American & 1{,}391 & 53{,}856  & 0.457 & 0.9286 [0.9031, 0.9418] & 0.1143 \\
          & Asian                    & 225     & 8{,}267   & 0.473 & 0.9227 [0.8945, 0.9433] & 0.1405 \\
          & Other                    & 468     & 20{,}255  & 0.513 & 0.9153 [0.8831, 0.9369] & 0.1253 \\
          & White                    & 3{,}761 & 160{,}994 & 0.522 & 0.8917 [0.8733, 0.9061] & 0.1379 \\
          & Hispanic / Latino        & 511     & 18{,}786  & 0.429 & 0.8222 [0.7735, 0.9229] & 0.1290 \\
\bottomrule
\end{tabular}
\end{table}

Twenty of the twenty-one
pairwise interval comparisons within attributes overlap; the exception
is the 40--54 versus 85+ age bands. No comparison by sex or
self-reported ethnicity separates. Interval overlap is descriptive and
is not a formal test of equal performance.
Excluding the Hispanic and Latino group discussed below,
subgroup AUROC is correlated with subgroup event rate at $-0.78$.
Across the age bands, event rate rises from 0.265 to 0.599 while AUROC
falls from 0.9277 to 0.8684. This association is consistent with
differences in case mix but does not establish that case mix causes the
performance differences.
Expected calibration error rises from 0.0962 in the youngest
band to 0.1728 in the oldest under the sampled evaluation protocol with ECG exclusions,
so the older bands also show larger calibration error in this
analysis.
The Hispanic and Latino subgroup has the lowest point estimate
(0.8222) and a wide confidence interval spanning 0.1494. That interval
contains the point estimate of every other ethnicity group, so the
data do not support a reliable conclusion of differential performance.
Its result also does not follow the same event-rate pattern as the
other ethnicity strata. Whether this reflects a genuine difference or
sampling imprecision cannot be settled here, and we flag it for
further evaluation before deployment.

\section{Clinical Prompt Design and Representative Examples}
\label{app:prompt}
Each training and inference instance is presented to the language model as a structured clinical question-answering prompt. The prompt opens with modality placeholder tokens, each replaced at forward time by an $8 \times 2{,}560$ soft-token matrix from the corresponding projector. It then provides a natural-language role description, a patient summary comprising demographic information, a compact time-series of vital sign measurements recorded in the 96-hour pre-arrival window, a stratified abnormality summary of all 470 structured clinical features, and a prior ECG history listing. The prompt closes with a binary prediction question instructing the model to answer with Yes or No.
When a patient has no prior ECG acquisitions within the 96-hour window, the prompt contains four modality tokens. When prior recordings exist, a fifth token carrying the aggregated longitudinal ECG representation is inserted, resulting in a five-token prompt variant. The two boxes below illustrate representative prompts from the test set.
\subsection{Box 1: Positive Case Without Prior ECG History}
\begin{tcolorbox}[
  colframe=black,
  colback=white,
  title=\footnotesize{74-year-old male $\cdot$ 24-hour mortality $\cdot$ Ground truth: Yes}
]
\scriptsize
\texttt{[EHR] [Retrieved EHR] [ECG] [Retrieved ECG]}
\medskip
You are a critical care physician with access to the patient's complete clinical record, current 12-lead ECG waveform, and Electronic Health Record (EHR). Your task is to predict whether a patient will experience mortality based on the ECG, the full 470-feature EHR summary, the vital signs time-series, and population evidence retrieved from similar patients.
\medskip
Patient: 74-year-old, Black/African, male.\\
Biometrics: not recorded.
\medskip
CRITICAL FINDINGS:\\
- Base excess: low at $-$20\,mmol/L (normal $-$2.0--3.0\,mmol/L).\\
- Glucose: critically elevated at 274\,mg/dL (normal 70.0--100.0\,mg/dL; 307$\rightarrow$274 $\downarrow$, slope $-$0.107/hr).\\
- INR: elevated at 1.3 (normal $<$1.1).\\
- Lactate: critically elevated at 10\,mmol/L (normal $<$2.0\,mmol/L; 9.9$\rightarrow$10 $\uparrow$, slope +0.0101/hr).\\
- pCO\textsubscript{2}: critically elevated at 64\,mmHg (normal 35.0--45.0\,mmHg).\\
- PT: elevated at 15.1\,sec (normal 11.0--13.5\,sec).\\
- PTT: critically elevated at 60.5\,sec (normal 25.0--35.0\,sec).\\
- WBC: critically elevated at 15.6$\times$10\textsuperscript{3}/\textmu L (normal 4.5--11.0$\times$10\textsuperscript{3}/\textmu L).
\medskip
WITHIN NORMAL LIMITS:\\
Creatinine; Fibrinogen; Ionized calcium; Hematocrit; Hemoglobin; Platelets; RDW; RBC; BUN.
\medskip
Prior ECG History (within 96h):\\
No prior ECG recordings identified in the 96h window.
\medskip
Will the patient die within 24 hours? Answer strictly with Yes or No.
\end{tcolorbox}
\subsection*{Box 2: Positive Case With Prior ECG History}
\begin{tcolorbox}[
  colframe=black,
  colback=white,
  title=\footnotesize{80-year-old female $\cdot$ ESI Level 3 $\cdot$ Two prior ECGs $\cdot$ 24-hour mortality $\cdot$ Ground truth: Yes}
]
\scriptsize
\texttt{[EHR] [Retrieved EHR] [ECG History] [ECG] [Retrieved ECG]}
\medskip
You are a critical care physician with access to the patient's complete clinical record, current 12-lead ECG waveform, and Electronic Health Record (EHR). Your task is to predict whether a patient will experience mortality based on the ECG, the full 470-feature EHR summary, the vital signs time-series, and population evidence retrieved from similar patients.
\medskip
Patient: 80-year-old, White, female. Acuity: ESI Level 3 (Urgent).\\
Biometrics: not recorded.
\medskip
CRITICAL FINDINGS:\\
- AST: elevated at 44\,U/L (normal 10.0--40.0\,U/L).\\
- Creatinine: critically elevated at 2\,mg/dL (normal 0.6--1.2\,mg/dL).\\
- Eosinophils: critically low at 0\% (normal 1.0--4.0\%).\\
- Glucose: elevated at 125\,mg/dL (normal 70.0--100.0\,mg/dL).\\
- Lactate: critically elevated at 5.1\,mmol/L (normal $<$2.0\,mmol/L).\\
- Lymphocytes: low at 14\% (normal 20.0--40.0\%).\\
- Neutrophils: elevated at 85\% (normal 50.0--70.0\%).\\
- Phosphate: elevated at 5.5\,mg/dL (normal 2.5--4.5\,mg/dL).\\
- Potassium: elevated at 5.2\,mEq/L (normal 3.5--5.0\,mEq/L).\\
- Troponin T: critically elevated at 0.11\,ng/mL (normal $<$0.01\,ng/mL).\\
- BUN: critically elevated at 39\,mg/dL (normal 7.0--20.0\,mg/dL).
\medskip
WITHIN NORMAL LIMITS:\\
ALT; Albumin; Alkaline phosphatase; Bands; Basophils; Bicarbonate; Total bilirubin; Total calcium; Chloride; Hematocrit; Hemoglobin; Magnesium; Platelets; RDW; RBC; Sodium; WBC.
\medskip
Prior ECG History (within 96h):\\
$\bullet$ 0.6h before current acquisition\\
$\bullet$ 0.4h before current acquisition
\medskip
Will the patient die within 24 hours? Answer strictly with Yes or No.
\end{tcolorbox}
\section{Training Data Construction and Prompt Generation}
\label{app:data_construction}
Each training and evaluation instance presented to PRISM is a structured clinical question-answering record derived from the MDS-ED benchmark~\citep{lopezalcaraz2025mdsed}. This appendix describes the end-to-end pipeline that transforms a raw MDS-ED task entry into the enriched prompt consumed by the MedGemma-4B backbone. The pipeline bridges the tabular MDS-ED data and the soft-token multimodal prompt format described in Sections~\ref{sec:memory_bank} and~\ref{sec:training}; representative prompt instantiations are shown in Appendix~\ref{app:prompt}.
\subsection{Source Data and Task Records}
The 1{,}443 binary prediction tasks are sourced from the MDS-ED benchmark, which pairs each clinical encounter with an ordered set of question-and-answer records of the form \emph{``Will the patient [outcome]? Answer: Yes / No''}. Each record contains a unique encounter identifier that maps to (i)~a row in the MDS-ED feature CSV providing all 470 structured EHR features, (ii)~a raw 12-lead ECG waveform path in MIMIC-IV-ECG, and (iii)~a longitudinal patient entry in the 96-hour time-series store containing timestamped vital-sign measurements and a list of prior ECG acquisitions. No additional outcome annotation was performed; ground-truth binary labels are taken verbatim from the published benchmark to preserve strict comparability with all previously reported baselines on identical data partitions.
\subsection{Population Statistics and Z-Score Normalization}
\label{sec:stats_norm}
Before any prompt is constructed, per-feature population statistics (mean and standard deviation) are computed over all numeric EHR columns from the 116{,}186 training-fold encounters (stratification folds~0--17 as defined by the MDS-ED protocol~\citep{lopezalcaraz2025mdsed}). These statistics serve two distinct roles. First, training-fold column medians are used for missing-value imputation in the FT-Transformer input vector (Section~\ref{sec:ehr_enc}), preventing evaluation-fold leakage into the structured encoder. Second, they define the reference distribution for z-score-based abnormality flagging in the text prompt: a measurement is designated \textsc{critical} when it either falls outside the published clinical reference interval (Table~\ref{tab:refranges}) or yields a population z-score of magnitude greater than 2.0, whichever criterion is the more inclusive. Severity is further stratified as \emph{elevated} or \emph{low} when the fractional deviation from the nearest bound is at most 30\%, and as \emph{critically elevated} or \emph{critically low} when the deviation exceeds 30\%.
\begin{table}[t]
\centering
\small
\setlength{\tabcolsep}{4pt}
\renewcommand{\arraystretch}{1.15}
\caption{Clinical reference intervals used for EHR abnormality flagging. Open bounds are indicated by a dash. Applied in conjunction with a population z-score threshold of $|z|>2.0$ (Section~\ref{sec:stats_norm}).}
\label{tab:refranges}
\begin{tabular}{llccc|llccc}
\toprule
\textbf{Cat.} & \textbf{Analyte} & \textbf{Lo} & \textbf{Hi} & \textbf{Unit} &
\textbf{Cat.} & \textbf{Analyte} & \textbf{Lo} & \textbf{Hi} & \textbf{Unit} \\
\midrule
Vitals & Temperature & 36.0 & 38.0 & $^\circ$C &
Vitals & Systolic BP & 90 & 140 & mmHg \\
 & Heart rate & 60 & 100 & bpm &
 & Diastolic BP & 60 & 90 & mmHg \\
 & Resp.\ rate & 12 & 20 & br/min &
 & O$_2$ sat. & 95 & -- & \% \\
\midrule
Hem. & WBC & 4.5 & 11.0 & $10^3$/\textmu L &
Hem. & Neutrophils & 50 & 70 & \% \\
 & hemoglobin & 12.0 & 17.5 & g/dL &
 & Lymphocytes & 20 & 40 & \% \\
 & hematocrit & 36 & 50 & \% &
 & Eosinophils & 1.0 & 4.0 & \% \\
 & Platelets & 150 & 400 & $10^3$/\textmu L &
 & Basophils & 0 & 1.0 & \% \\
 & RBC & 3.8 & 5.8 & M/\textmu L &
 & Bands & -- & 5.0 & \% \\
 & RDW & 11.5 & 14.5 & \% &
 & & & & \\
\midrule
Coag. & INR & -- & 1.1 & -- &
Coag. & PTT & 25 & 35 & sec \\
 & PT & 11.0 & 13.5 & sec &
 & Fibrinogen & 200 & 400 & mg/dL \\
\midrule
Chem. & Sodium & 136 & 145 & mEq/L &
Chem. & Potassium & 3.5 & 5.0 & mEq/L \\
 & Chloride & 98 & 106 & mEq/L &
 & Bicarbonate & 22 & 29 & mEq/L \\
 & BUN & 7.0 & 20 & mg/dL &
 & Creatinine & 0.6 & 1.2 & mg/dL \\
 & Glucose & 70 & 100 & mg/dL &
 & Total calcium & 8.5 & 10.5 & mg/dL \\
 & Phosphate & 2.5 & 4.5 & mg/dL &
 & Magnesium & 1.7 & 2.2 & mg/dL \\
 & Albumin & 3.4 & 5.4 & g/dL &
 & & & & \\
\midrule
Hepatic & ALT & 7.0 & 56 & U/L &
Hepatic & AST & 10 & 40 & U/L \\
 & Alk.\ phos. & 44 & 147 & U/L &
 & Total bili. & -- & 1.2 & mg/dL \\
\midrule
Cardiac & Troponin T & -- & 0.01 & ng/mL &
Cardiac & Total CK & -- & 200 & U/L \\
 & CK-MB & -- & 5.0 & ng/mL &
 & & & & \\
\midrule
Metabolic & Lactate & -- & 2.0 & mmol/L &
Metabolic & pCO$_2$ & 35 & 45 & mmHg \\
 & Base excess & $-$2 & 2.0 & mmol/L &
 & pH & 7.35 & 7.45 & -- \\
 & Ion.\ Ca & 1.12 & 1.32 & mmol/L &
 & & & & \\
\bottomrule
\end{tabular}
\end{table}
\subsection{Language Model Prompt}
The prompt for each instance is assembled from five ordered components.
\paragraph{Modality token prefix.}
Between four and five placeholder tokens are prepended to the prompt, one per active modality: the current-visit ECG waveform (\texttt{<ecg>}), the FT-Transformer EHR encoding (\texttt{<ehr>}), the retrieved EHR population context (\texttt{<retrieved\_ehr>}), the aggregated longitudinal ECG history (\texttt{<ecg\_history>}, inserted only when the patient has at least one prior acquisition within the 96-hour window), and the retrieved ECG population context (\texttt{<retrieved\_ecg>}). At forward time, each placeholder is replaced in-place by an $8 \times 2{,}560$ soft-token matrix from the corresponding projector (Table~\ref{tab:proj}). Encounters without prior ECG recordings receive the four-token variant; all others receive the five-token variant.
\paragraph{Role and task block.}
A fixed natural-language role description identifies the model as a critical care physician and specifies the prediction target for the current instance (e.g., \emph{``Will the patient be diagnosed with `Atrial fibrillation'?''}). This framing is derived verbatim from the binary question string in the MDS-ED task record and is standardized uniformly across all 1{,}443 tasks.
\paragraph{Patient summary block.}
The patient summary comprises four programmatically generated sub-sections: (i)~a demographics and biometrics header containing age, sex, self-reported ethnicity, body-mass index, weight, height, and the Emergency Severity Index acuity level when recorded; (ii)~a time-stamped vital-sign table spanning up to ten representative rows from the 96-hour window (when more than ten measurements are available, the earliest and nine most recent observations are retained); (iii)~a stratified EHR abnormality summary classifying all 470 features under \textsc{critical findings} or \textsc{within normal limits} according to the criteria in Section~\ref{sec:stats_norm}, including observed value, unit, reference interval, and trend direction; and (iv)~a listing of the number and relative timestamps (in hours before the index ECG) of prior ECG acquisitions within the 96-hour window.
\subsection{Memory Bank Lookup Integration}
After Stage~1 training, the FT-Transformer and D-BETA encoder produce fixed-length embeddings for all 116{,}186 training-fold encounters. These embeddings populate two independent nearest-neighbor indices, one per modality, as described in Appendix~\ref{app:memory_bank}. During the prompt-construction pass that precedes Stage~2 training and evaluation, each encounter is assigned its top-$k$ nearest neighbors from each index via a single lookup against pre-computed neighbor tables materialized immediately after Stage~1. The resulting neighbor row indices are embedded directly in the serialized data record so that the Stage~2 data loader can reconstruct the corresponding embedding vectors at batch time without incurring query latency during distributed GPU training.
\subsection{Train, Validation, and Test Handling}
Validation and test encounters (stratification folds~18 and~19, respectively) are processed through an identical pipeline, with one critical exception: their memory bank queries target the \emph{training-fold} index exclusively, ensuring that no ground-truth outcome information from held-out partitions enters the retrieved population context.
\section{Population Memory Bank: Retrieval Across Clinical Phenotypes}
\label{app:retrieval_phenotypes}
\begin{figure*}[h]
  \centering
  \captionsetup{font=scriptsize}
  \includegraphics[width=0.7\textwidth]{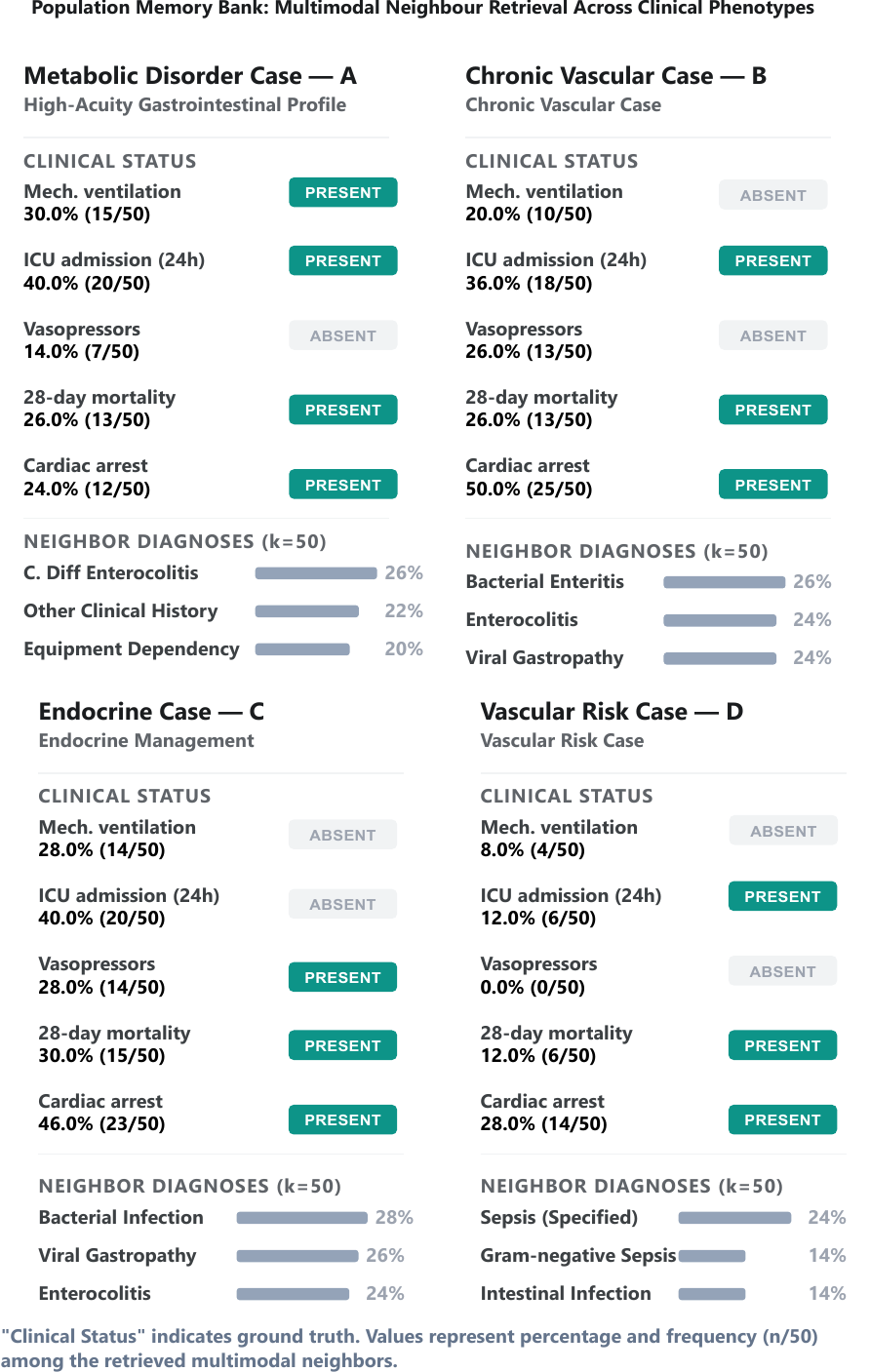}
  \caption{
    \textbf{Population memory bank retrieval across four clinical phenotypes (Cases A--D).}
    For each test encounter, the 50 nearest neighbors of the EHR index are used, so every percentage below is computed against a denominator of 50; clinical status (PRESENT/ABSENT) reflects the ground-truth outcome label of the index patient, and percentages report outcome prevalence among the 50 retrieved neighbors.
  }
  \label{fig:retrieval_phenotypes}
\end{figure*}
Figure~\ref{fig:retrieval_phenotypes} extends the single-encounter retrieval analysis of Section~\ref{sec:retrieval_analysis} to four clinically distinct test-set encounters, each representative of a different phenotypic category in the MDS-ED cohort. For each encounter, we take the 50 nearest neighbors of the EHR index, so all proportions below have a denominator of 50; ground-truth binary outcome labels and discharge diagnosis codes are aggregated by direct counting.
The four selected cases illustrate different local neighborhoods.
For example, the high-acuity metabolic case has neighbor prevalences of
30.0\%, 40.0\%, 26.0\%, and 24.0\% for mechanical ventilation, ICU
admission, 28-day mortality, and cardiac arrest, respectively, while
the chronic vascular case has 50.0\% cardiac-arrest prevalence among
its 50 EHR neighbors. The remaining cases show different diagnostic
and outcome compositions. These examples are intentionally selected to
span distinct phenotypes, and $k{=}50$ is used to make prevalence
estimates readable. They illustrate index geometry only and do not
establish cohort-level outcome concordance or the prediction mechanism.
\begingroup

\section{External and Temporal Evaluation Protocols}
\label{app:external_temporal_protocol}
\paragraph{MC-MED cohort and prompts.}
Table~\ref{tab:external_temporal_cohorts} reports the temporal and external cohorts. The MC-MED cohort contains 40{,}121 visits from 30{,}642 patients. The local memory bank contains the 35{,}818 training and validation visits. Patient-overlap checks using medical record numbers found zero shared patients in all three partition pairs: training--validation, training--test, and validation--test. Consequently, the local memory bank, constructed from training and validation visits, contains no test patients. Training comprises 109{,}390 prompt records over 682 tasks, of which a stratified subset of 49{,}120 is used for fine-tuning. Validation contains 14{,}648 records over 314 tasks. The 13{,}254 test records cover 3{,}588 patients and 306 tasks; tasks enter the test prompts only if at least five positive examples are available.

\begin{table}[t]
\centering

\captionsetup{font=scriptsize}
\caption{\textbf{Cohort sizes for external and temporal evaluation.} Counts describe the underlying visit/encounter tables, rather than the sampled encounter--task records. Banks comprise training and validation rows. }
\label{tab:external_temporal_cohorts}
\small
\renewcommand{\arraystretch}{1.25}
\begin{tabular}{@{}llrr@{}}
\toprule
\textbf{Setting} & \textbf{Partition} & \textbf{Encounters/visits} & \textbf{Patients} \\
\midrule
Temporal & Training & 67{,}562 & 43{,}238 \\
 & Validation & 3{,}524 & 2{,}342 \\
 & Test & 1{,}435 & 1{,}002 \\
 & Memory bank & 71{,}086 & 45{,}580 \\
\midrule
MC-MED & Training & 31{,}239 & 22{,}908 \\
 & Validation & 4{,}579 & 3{,}982 \\
 & Test & 4{,}303 & 3{,}752 \\
 & Total cohort & 40{,}121 & 30{,}642 \\
 & Local memory bank & 35{,}818 & 26{,}890 \\
\bottomrule
\end{tabular}
\end{table}

\paragraph{MC-MED construction and local fine-tuning.}
Cohort construction reduces 118{,}385 source visits to 40{,}121 included visits. The available lead-II waveform is repeated across all 12 input channels for ECG encoding; this supplies a single measured lead rather than a diagnostic 12-lead recording. The structured-data mapping covers 43 of the 45 laboratory analytes. Local fine-tuning uses 49{,}120 stratified prompt records for two epochs. Training takes 2 hours 21 minutes on two H100 GPUs.

\paragraph{Eligible outcomes.}
Primary analyses retain tasks with at least five positive and five negative evaluation records. The temporal evaluation contains 841 tasks: 831 diagnosis tasks; mechanical ventilation and vasopressors; ICU admission within 24 hours and during the stay; and mortality within 1, 7, 28, 90, 180, and 365 days. The MC-MED evaluation contains 306 tasks: 297 diagnosis tasks; inotropes, mechanical ventilation, severe hypoxemia, and vasopressors; ICU admission within 24 hours; and in-hospital mortality and mortality within 1, 7, and 28 days. Group averages therefore summarize different eligible task sets across settings and are not direct estimates of a performance change on identical outcomes.

\paragraph{Prediction and aggregation.}
The PRISM rows in Tables~\ref{tab:mcmed_external} and~\ref{tab:temporal_generalization} use the generative pathway: softmax over the Yes/No answer-token logits at the final prompt position. The discriminative pathway instead uses a classification head on the mean-pooled prompt representation. Baselines produce a single score. AUROC uses average-rank tie handling. MC-MED tasks are evaluated on positives and an equally sized random negative sample. For MDS-ED, balancing is performed before ECG exclusions over the full test fold; the temporal evaluation then retains later-era records without resampling. Neither ECG exclusion nor temporal restriction preserves exact within-task balance. All rows within each table use identical retained evaluation records. Per-task AUROCs are averaged within each eligible task group; overall AUROC is the unweighted mean of the four group means.

\paragraph{Uncertainty.}
Bracketed values are 95\% percentile confidence intervals from 1{,}000 patient-level cluster-bootstrap resamples, using MIMIC-IV subject identifiers or MC-MED medical record numbers as the cluster unit. A common resample is used across compared runs, enabling paired contrasts. Fusion baselines use multiple seeds, whose scores are averaged within each resample. Each reported PRISM configuration uses a single training run. The gradient-boosting baseline reported in the MC-MED evaluation also uses a single run. These intervals characterize patient-sampling uncertainty and do not quantify PRISM's training-seed variability.

\paragraph{MC-MED retrieval controls.}
Table~\ref{tab:mcmed_retrieval_controls} reports the group-specific uncertainty for the retrieval controls in Table~\ref{tab:mcmed_external}. Both controls use the fine-tuned PRISM checkpoint and its generative pathway. Retrieval is perturbed at inference without retraining, with the evaluation records held fixed. The intervals describe each condition separately rather than the paired differences between conditions.

\begin{table}[t]
\centering

\captionsetup{font=scriptsize}
\caption{\textbf{MC-MED retrieval controls.} AUROC and 95\% patient-level percentile-bootstrap confidence intervals for inference-time perturbations of fine-tuned PRISM. Overall is the unweighted mean of the four task-group AUROCs.}
\label{tab:mcmed_retrieval_controls}
\small
\renewcommand{\arraystretch}{1.25}
\begin{tabular}{@{}lcc@{}}
\toprule
\textbf{Task group} & \textbf{Retrieval zeroed} & \textbf{Random retrieval} \\
\midrule
Diagnosis & 0.7920 [0.779, 0.807] & 0.7919 [0.778, 0.806] \\
Deterioration & 0.7389 [0.641, 0.831] & 0.7310 [0.628, 0.828] \\
ICU & 0.8681 [0.830, 0.903] & 0.8657 [0.826, 0.901] \\
Mortality & 0.9215 [0.887, 0.949] & 0.9236 [0.889, 0.950] \\
\midrule
Overall & 0.8301 [0.802, 0.857] & 0.8280 [0.799, 0.856] \\
\bottomrule
\end{tabular}
\end{table}

\paragraph{Source-table counts.}
The MDS-ED source tables used for the temporal evaluation contain 116{,}433 encounters from 63{,}929 patients in folds 0--17, 6{,}185 encounters from 3{,}525 patients in fold 18, and 6{,}439 encounters from 3{,}644 patients in fold 19. These source-table row counts are distinguished from the task-eligible counts reported for the original benchmark evaluation.
\endgroup

\begingroup

\section{Temporal Generalization}
\label{app:temporal_generalization}
We additionally evaluate PRISM under a temporal split, training on encounters assigned with certainty to 2016 or earlier and testing on encounters assigned with certainty to 2017 or later. Encounters with date intervals crossing the cutoff are excluded. Training and test patients are disjoint. Table~\ref{tab:temporal_generalization} compares PRISM with compact fusion baselines on the same evaluation records. Overall AUROC is the unweighted mean of the four task-group AUROCs.

\paragraph{Partitions and evaluation coverage.}
Training comprises folds 0--17 restricted to encounters whose real year is certainly at most 2016; validation comprises fold 18 under the same restriction. Test encounters are from fold 19 with real year certainly at least 2017. The memory bank contains the union of training and validation, totaling 71{,}086 encounters from 45{,}580 patients. No test patient appears in the training, validation, or bank sets. Table~\ref{tab:external_temporal_cohorts} gives the partition counts.

The temporal evaluation retains the later-era records from the original sampled test set without resampling. It contains 59{,}493 encounter--task records from 998 of the 1{,}002 test patients. Eight of the 83 encounters excluded for unavailable ECGs are from 2017 or later and account for the four later-era patients with no retained record. Balancing was performed over the full test fold, so the later-era subset is not exactly balanced within tasks. The primary analysis includes 841 tasks meeting the five-positive/five-negative threshold; 1{,}270 tasks have a defined AUROC under the broader sensitivity criterion. Training contains 2{,}642{,}334 prompt records, with stratified subsets of 704{,}811 for Stage~1 and 99{,}820 for Stage~2. Validation contains 126{,}544 prompt records. PRISM uses the generative prediction pathway, and uncertainty follows Appendix~\ref{app:external_temporal_protocol}.

\begin{table}[t]
\centering

\captionsetup{font=scriptsize}
\caption{\textbf{Temporal generalization to later encounters.} All entries are AUROC; PRISM uses the generative pathway. Brackets show reported 95\% patient-bootstrap confidence intervals. The later-era records are retained without rebalancing after the original test-set ECG exclusions. Zeroed and random retrieval are inference-time controls. Bold indicates the highest point estimate in each column, not statistical significance.}
\label{tab:temporal_generalization}
\small
\setlength{\tabcolsep}{3pt}
\renewcommand{\arraystretch}{1.25}
\resizebox{\textwidth}{!}{%
\begin{tabular}{@{}lccccc@{}}
\toprule
\textbf{Model / condition} & \textbf{Diagnosis} & \textbf{Deterioration} & \textbf{ICU} & \textbf{Mortality} & \textbf{Overall AUROC} \\
\midrule
Fusion MLP & 0.6617 & 0.8049 & 0.7858 & 0.8153 & 0.7669 [0.730, 0.801] \\
Fusion MLP + retrieval & 0.6488 & 0.7941 & 0.7798 & 0.7935 & 0.7541 [0.716, 0.788] \\
PRISM & \textbf{0.8119} & 0.8501 & \textbf{0.8726} & \textbf{0.9004} & \textbf{0.8587} [0.821, 0.892] \\
PRISM, retrieval zeroed & 0.8064 & 0.8431 & 0.8665 & 0.8957 & 0.8529 [0.815, 0.887] \\
PRISM, random neighbors & 0.8099 & \textbf{0.8525} & 0.8717 & 0.8988 & 0.8582 [0.821, 0.891] \\
\bottomrule
\end{tabular}}
\end{table}

PRISM reaches an overall AUROC of 0.8587, compared with 0.7669 for the compact fusion MLP and 0.7541 for fusion with retrieval. These correspond to point-estimate differences of 9.18 and 10.46 AUROC points, respectively. PRISM also exceeds both compact fusion baselines in each task group. These comparisons concern the complete model configurations rather than isolating the language-model backbone.

Zeroing retrieval reduces PRISM's overall AUROC by 0.58 points, but genuine retrieval exceeds random neighbors by only 0.05 points. These results distinguish sensitivity to removing retrieval context from the benefit of similarity-based neighbor selection: the latter is small in this temporal setting. Adding retrieval also does not improve the compact fusion baseline. We interpret these as setting-dependent effects and do not infer statistical significance from the separate model intervals.

\endgroup
\end{document}